\documentclass{article}

\usepackage[preprint]{neurips_2026}

\usepackage[utf8]{inputenc}
\usepackage[T1]{fontenc}
\usepackage[hidelinks]{hyperref}
\usepackage{url}
\usepackage{booktabs}
\usepackage{amsfonts}
\usepackage{amsmath}
\usepackage{nicefrac}
\usepackage{microtype}
\usepackage{xcolor}
\usepackage{graphicx}
\usepackage{subcaption}
\usepackage{float}
\usepackage{soul}

\graphicspath{{official_figures/}{figures/}}

\title{Frontier Coding Agents Use Metaprogramming to Adapt to Unfamiliar Programming Languages}

\author{%
  Aman Sharma\thanks{Primary author. Correspondence: \texttt{aman.sharma@lossfunk.com}.} \\
  Lossfunk \\
  \texttt{aman.sharma@lossfunk.com} \\
  \And
  Sushrut Thorat\addtocounter{footnote}{1}\thanks{Equal advising.} \\
  Lossfunk \\
  \texttt{sushrut.thorat@lossfunk.com} \\
  \And
  Paras Chopra\footnotemark[3] \\
  Lossfunk \\
  \texttt{paras@lossfunk.com} \\
}

\begin{document}

\begin{center}
\includegraphics[height=2.0cm]{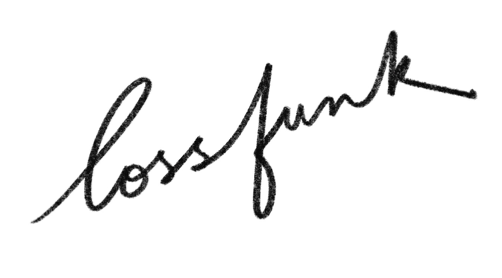}
\end{center}
\vspace{-0.5em}

\maketitle

\begin{abstract}
LLM-based coding agents are usually evaluated in familiar software
settings: mainstream languages, common libraries, and public
repositories. These benchmarks remain important, but they can hide how
agents behave when the language itself is unfamiliar. We evaluate six
contemporary coding agents on four esoteric programming languages
using a sequential setup with file editing, local execution, and
hidden-test grading. Our protocol exposes capability differences
between these agents that mainstream coding and agentic benchmarks such as SWE-Bench Verified and Terminal-Bench 2.0 compress into much
narrower bands. We observe that the strongest agents, Claude Opus 4.6 and GPT-5.4 xhigh,
often avoid writing the target language directly. On Brainfuck and
Befunge-98, they write Python programs that generate target-language code
and debug those generators locally. Forbidding this metaprogramming strategy
causes large performance drops. Text guidance distilled from this strategy
does not materially improve weaker agents. In contrast, Opus-derived Python
helper code for building generators, with no solved benchmark programs or
hidden-test answers, sharply improves Sonnet 4.6 and GPT-5.4 mini on the
same problems, while Haiku 4.5 remains low. More interpreter calls and output tokens improve stronger agents but leave
weaker agents near their original performance, indicating that these resources
amplify useful strategies rather than create them. Together, these results show that \textbf{strong coding agents adapt to unfamiliar
languages by using tools, feedback, and workspace state to build a working model
of the target language}. Metaprogramming is the clearest case, but the broader
gap is constructing and debugging a strategy that works under the target
language's rules.
\end{abstract}

\section{Introduction}

Coding is one of the central applications of large language model (LLM) agents. Most prominent benchmarks for coding agents evaluate them in familiar software ecosystems: mainstream languages, common libraries, and public open-source repositories. SWE-Bench Verified~\citep{jimenez2024swebench} is a canonical example, testing agents on real GitHub issues from widely used Python projects. These benchmarks measure important progress on realistic software-engineering tasks, but they also test settings where frontier models have extensive prior exposure to the relevant syntax, APIs, libraries, coding patterns, and repository structure.
A complementary question is how the same agents behave when the programming language itself is unfamiliar: when the agent must figure out how to write, run, debug, and revise code in a language whose syntax and execution rules are not already familiar. This setting has received comparatively less attention in agentic coding evaluation, despite its practical relevance. Production systems often require models to work with internal domain-specific languages, proprietary configuration files, generated APIs, and local tool conventions that are absent from public corpora or differ from standard programming environments. In such settings, success depends less on recalling familiar code patterns and more on building a working understanding of the target interface during the session.

\begin{figure}[t]
  \centering
  \includegraphics[width=0.9\linewidth]{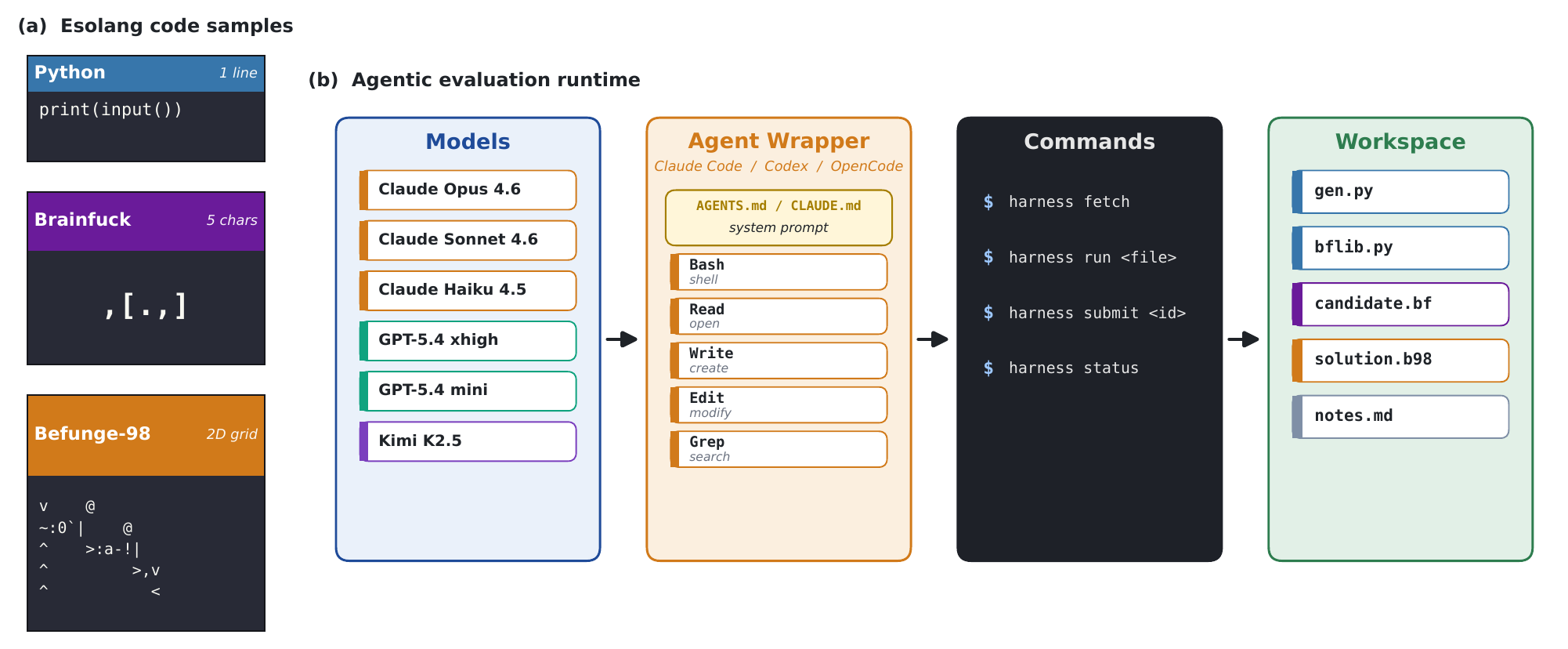}
  \caption{\textbf{Task substrate and agentic runtime.}
  (a) The same simple input-and-print task in Python, Brainfuck, and Befunge-98 shows how different esolang code looks from ordinary code. (b) Each model runs
  in a coding harness (Claude Code, Codex, or OpenCode) with file editing,
  shell access, benchmark commands, and a persistent workspace for local
  execution and hidden-test submission.}
  \label{fig:tasks-harness}
\end{figure}

To study how contemporary LLM-based coding agents behave when the programming language itself is unfamiliar, we use languages from EsoLang-Bench~\citep{esolangbench2026}. These esoteric languages are not realistic production targets; they are controlled proxies for unfamiliar executable interfaces. For example, Brainfuck is a minimal pointer-machine language, and Befunge-98 introduces two-dimensional control flow over a stack-based grid (Figure~\ref{fig:tasks-harness}a). This makes them useful for testing whether agents can learn an unfamiliar language well enough during a session to write, run, debug, and improve working programs.

We therefore build an agentic evaluation pipeline around EsoLang-Bench and use it to compare a capability ladder of six contemporary LLM-based coding agents under a common tool-use protocol: Claude Opus 4.6, Sonnet 4.6, and Haiku 4.5; GPT-5.4 xhigh and GPT-5.4 mini (xhigh and medium reasoning effort, respectively); and Kimi K2.5 (Figure~\ref{fig:tasks-harness}). Each agent works in a persistent workspace where it can edit files, run code locally, and submit final answers to hidden tests. The evaluation therefore tests an interactive problem-solving process, not a single code completion. We analyze final pass rates together with agent logs and targeted interventions, allowing us to ask which agents succeed and how they adapt.

\paragraph{Central observations:}
\begin{enumerate}

  \item \textbf{Unfamiliar-language evaluation separates agents that look similar on mainstream coding benchmarks.}
  Under our EsoLang-Bench protocol, where the target language must be worked out within the session, these agents are separated over a much wider range than on mainstream coding benchmarks such as SWE-Bench Verified and Terminal-Bench 2.0, exposing capability differences that those benchmarks compress into narrower bands (Table~\ref{tab:cliff-spread}).

  \item \textbf{The strongest agents use metaprogramming.}
  They write Python generators that emit target-esolang programs, reuse helpers across problems, and test locally before submission. This emerges without language-specific prompting. Forbidding metaprogramming on Brainfuck and Befunge-98 drops performance by tens of percentage points.

    \item \textbf{Strategy transfer works through executable scaffolds, but not with distilled written strategies.}
  A textual description of the strategy does not close the gap, but providing the implementation of that strategy as a reference library of working Python generators substantially improves Sonnet 4.6 and GPT-5.4 mini. Haiku 4.5 remains low, showing that some agents still cannot compose the provided machinery into working solutions.

  \item \textbf{Extra inference-time resources help only when agents can use them.}
  More interpreter calls and output tokens improve stronger agents but leave weaker agents near their original performance, indicating that resources amplify useful strategies rather than create them.

\end{enumerate}


\section{Experimental setup}
\label{sec:eval}

\paragraph{Task substrate.}
We evaluate on EsoLang-Bench~\citep{esolangbench2026}, using the original
80-problem sequences for Brainfuck, Befunge-98, Whitespace, and Shakespeare.
EsoLang-Bench includes a fifth language, Unlambda, which we exclude because
its interpreter made local execution substantially slower in our agentic
setting; including it would make wall-clock runtime depend heavily on
interpreter latency rather than adaptation behavior. The problems themselves
are short standard programming tasks (echoing input, sorting integers,
GCD/LCM, and similar list- and number-manipulation tasks across the four
difficulty tiers; the full per-tier task list is in
Appendix~\ref{app:task-details}). The original EsoLang-Bench evaluation reports near-ceiling
performance on these same problem statements when models answer in Python
or JavaScript, so here the difficulty is primarily expressing, implementing,
and debugging solutions in an unfamiliar target language.

\paragraph{Agentic protocol.}
We use the same four-language task substrate as EsoLang-Bench, but evaluate
each model as an interactive coding agent rather than a one-pass generator.
Each model$\times$language run is one sequential session over all $80$ problems
for that language. Problems are fetched in fixed forward order. For each
problem, the agent receives the statement, edits files in an isolated
workspace, runs candidates locally, and may make up to three hidden
submissions. A problem is finalized when one submission passes all six hidden
tests or when the three submissions are exhausted; finalized problems are not
revisited.

\begin{figure}[t]
  \centering
  \includegraphics[width=0.75\linewidth]{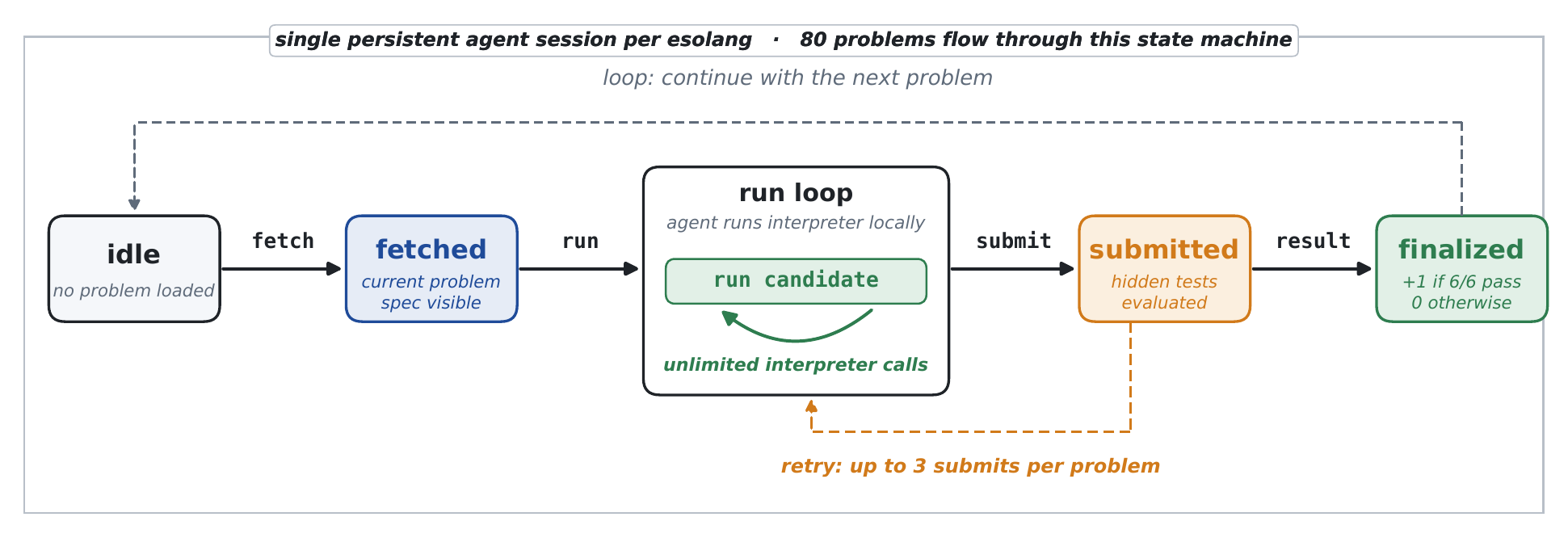}
  \caption{\textbf{Per-problem state machine under the primary protocol.}
  Each model--language run is a fixed forward session over $80$ problems.
  For each problem, the agent fetches the specification, edits and
  executes candidate programs locally, and makes up to three hidden
  submissions. Hidden submissions return only aggregate hidden-test
  feedback; finalized problems are not revisited.}
  \label{fig:methodology-pipeline}
\end{figure}

Local interpreter calls expose ordinary execution feedback such as stdout,
stderr, and runtime errors. Hidden submissions return only the aggregate
number of private tests passed, not the private inputs, expected outputs, or
per-test diagnostics. Figure~\ref{fig:methodology-pipeline} summarizes the state machine.
The primary protocol uses $80$ problems per language, six hidden tests per
problem, up to three hidden submissions, unlimited local interpreter calls,
a $32$k-token output budget per assistant turn, and isolated workspaces;
all parameters are summarized in Appendix Table~\ref{tab:setup}.

\paragraph{Models and harnesses.}
We evaluate deployed coding agents rather than bare models. Claude Opus
4.6, Claude Sonnet 4.6, and Claude Haiku 4.5 run under Claude
Code~\citep{anthropic_models_2026}; GPT-5.4 xhigh and GPT-5.4 mini run
under Codex~\citep{openai_gpt54_2026,openai_gpt54mini_2026}; and Kimi
K2.5 runs under OpenCode~\citep{moonshot_kimi_k25_2026}. This
model$\times$harness pairing is part of what we evaluate, because tool
mediation, file editing, shell access, and workspace management are part
of deployed coding-agent systems. Per-agent API endpoints, model
identifiers, sampling settings, and harness invocations are documented in
Appendix~\ref{app:experimental-details}.

Every agent receives the same benchmark-facing operations and the same
per-language system prompt (a simple task prompt for benchmarking,
with no problem-specific guidance, no solved examples, and no
hidden-test material); the system prompt and the per-condition
deviations from this primary prompt are in
Appendix~\ref{app:system-prompts}. As a cross-harness check, we re-ran
Opus 4.6 and GPT-5.4 xhigh under OpenCode on Brainfuck and Befunge-98;
we observed similar performance and the qualitative ordering is unchanged
(Appendix~\ref{app:cross-harness}).

\paragraph{Logging and behavioral measurements.}
For each run, we log problem fetches, shell commands, local interpreter
calls, hidden submissions, file edits, generated files, command outputs, and
final workspace state. We use \emph{metaprogramming} to mean that the agent
writes a program in a familiar host language, such as Python, JavaScript, or
Rust, whose output is source code in the target esolang. This differs from
direct authoring, where the agent edits the target esolang source itself. A
helper file is reusable if it persists in the workspace and is called,
imported, copied, or modified across multiple problems in the same session.
These labels describe behavior only; scoring depends solely on hidden-test
success.

\paragraph{Scoring and reporting.}
A problem is counted as solved if and only if one of the agent's
submissions passes all six private hidden tests for that problem; a
submission that passes only a subset of the hidden tests counts as a
failure for that submission, with no partial credit, and we do not
aggregate hidden-test passes across the up-to-three submissions
allowed per problem. Primary scores are solved problems out of $80$
for each model$\times$language run.

For each model$\times$language cell, we run three independent sessions
under the same task order and protocol. We report the first session's
solved count as the headline value in
Table~\ref{tab:main-results} and Table~\ref{tab:cliff-spread}, with
Wilson $95\%$ binomial confidence
intervals~\citep{wilson1927probable} computed over its 80 per-problem
outcomes ($k/80$ per-language and $k/320$ pooled). The remaining two
sessions serve as session-to-session sanity checks; per-session
counts for all three sessions are tabulated in
Appendix~\ref{app:per-session-results}. Full reporting and
confidence-interval details are in Appendices~\ref{app:reporting},
\ref{app:robustness}, and~\ref{app:wilson-asym}.

\paragraph{Diagnostic protocol variants.}

All headline results use the primary protocol above. We use protocol variants only for targeted diagnostic experiments. In Section~\ref{sec:abl-meta}, we restrict metaprogramming by requiring agents to author Brainfuck and Befunge-98 directly, without using a host-language generator. In Section~\ref{sec:distillation}, we test strategy transfer by giving weaker agents either text guidance distilled from Claude Opus 4.6's traces or a small reference library of working generator code, with no solved benchmark problems or hidden-test answers included. In Section~\ref{sec:resources}, we vary local interpreter-call budgets and output-token budgets while holding the task substrate fixed.

These variants are used to explain the performance gaps observed under the primary protocol, not to define the main score. Unless explicitly stated otherwise, model prompts, task order, hidden tests, submission limits, and scoring rules remain unchanged from the primary protocol.

\begin{table}[tbp]
  \caption{\textbf{Main EsoLang-Bench results under the primary protocol.}
  EsoLang cells report Session 1 percentage solved (the headline
  session; two further sessions per cell are reported in
  Appendix~\ref{app:per-session-results}) with Wilson 95\% binomial
  confidence intervals as subscripts. The \textbf{Mean} column
  averages the four esolang scores.}
  \label{tab:main-results}
  \centering
  \small
  \setlength{\tabcolsep}{4pt}
  \begin{tabular}{@{}lccccc@{}}
    \toprule
    Agent & WS & Sh & B98 & BF & \textbf{Mean} \\
    \midrule
    Kimi K2.5
      & $31.3_{\scriptstyle \pm 10.8}$ & $2.5_{\scriptstyle +6.2}$ & $6.3_{\scriptstyle +7.6}$ & $5.0_{\scriptstyle +7.2}$
      & $\mathbf{11.3_{\scriptstyle \pm 3.9}}$ \\
    Haiku 4.5
      & $81.3_{\scriptstyle \pm 9.9}$ & $7.5_{\scriptstyle +7.9}$ & $5.0_{\scriptstyle +7.2}$ & $5.0_{\scriptstyle +7.2}$
      & $\mathbf{24.7_{\scriptstyle \pm 5.0}}$ \\
    Sonnet 4.6
      & $100.0_{\scriptstyle -4.6}$ & $70.0_{\scriptstyle \pm 10.8}$ & $80.0_{\scriptstyle \pm 10.0}$ & $15.0_{\scriptstyle \pm 9.4}$
      & $\mathbf{66.3_{\scriptstyle \pm 5.3}}$ \\
    Opus 4.6
      & $100.0_{\scriptstyle -4.6}$ & $87.5_{\scriptstyle \pm 9.0}$ & $80.0_{\scriptstyle \pm 10.0}$ & $80.0_{\scriptstyle \pm 10.0}$
      & $\mathbf{86.9_{\scriptstyle \pm 4.1}}$ \\
    GPT-5.4 mini
      & $88.8_{\scriptstyle \pm 8.8}$ & $21.3_{\scriptstyle \pm 10.2}$ & $13.8_{\scriptstyle \pm 9.2}$ & $6.3_{\scriptstyle +7.6}$
      & $\mathbf{32.5_{\scriptstyle \pm 5.3}}$ \\
    GPT-5.4 xhigh
      & $100.0_{\scriptstyle -4.6}$ & $100.0_{\scriptstyle -4.6}$ & $100.0_{\scriptstyle -4.6}$ & $98.8_{\scriptstyle -5.5}$
      & $\mathbf{99.7_{\scriptstyle -1.4}}$ \\
    \bottomrule
  \end{tabular}

  \vspace{0.25em}
  {\footnotesize
  BF=Brainfuck, B98=Befunge-98, WS=Whitespace, Sh=Shakespeare.
  Subscripts denote Wilson 95\% intervals: $-x$ for ceiling cells, $+x$
  for near-floor cells, and $\pm x$ for interior cells. Raw counts
  are in Appendix~\ref{app:main-results-raw}, per-session counts in
  Appendix~\ref{app:per-session-results}, and full asymmetric
  intervals in Appendix~\ref{app:wilson-asym}.}
\end{table}

\section{Results}
\label{sec:results}

\subsection{Unfamiliar-language evaluation sharply separates contemporary coding agents}
\label{sec:cliff}

We first ask how contemporary agentic models fare under unfamiliar-language evaluation. Table~\ref{tab:main-results} reports
reports per-language EsoLang-Bench scores for the six evaluated agents under
the primary protocol, with Wilson $95\%$ binomial confidence intervals
(Section~\ref{sec:eval}, Appendix~\ref{app:wilson-asym}). We observe a large performance spread between the agents. The separation is not driven by a single uniformly hard language. Per-language ranges are $5.0$--$98.8$ on Brainfuck, $5.0$--$100$ on Befunge-98, $31.3$--$100$ on Whitespace, and $2.5$--$100$ on Shakespeare. Whitespace is near-ceiling for several agents, whereas Brainfuck and Befunge-98 expose large separations. What these two languages share is that they are low-ecosystem targets whose syntax and idioms are far from mainstream software work; the spread is therefore informative about within-session adaptation to an unfamiliar executable interface.

Comparing the same six agents across mainstream coding benchmarks reveals a much smaller spread. Table~\ref{tab:cliff-spread} reports the spread and standard deviation across three mainstream coding benchmarks (SWE-Bench Verified, Terminal-Bench 2.0, and LiveCodeBench v6) together with the EsoLang-Bench mean; EsoLang-Bench produces both the widest spread and the largest SD of the four. Figure~\ref{fig:cliff-scatter} in the appendix visualizes the SWE-V versus EsoLang-Bench cell as a scatter plot. This result shows that unfamiliar-language evaluation exposes capability differences that these mainstream benchmarks compress.

\begin{table}[tbp]
  \caption{\textbf{Mainstream coding scores cluster while
  unfamiliar-language scores separate.} Six contemporary coding
  agents on three mainstream coding benchmarks and the EsoLang-Bench
  four-language mean. Spread is best minus worst across the six
  agents; SD is the sample standard deviation ($n{=}6$, $n{-}1$
  denominator). The EsoLang SD ($36.0$) is roughly $12\times$
  SWE-Bench Verified's ($2.9$) and $2\times$ LiveCodeBench v6's
  ($17.2$). \textsuperscript{$\dagger$}\,vendor-reported;
  \textsuperscript{$\ast$}\,Vals.ai. Per-agent sourcing in Appendix
  Table~\ref{tab:swev-sourcing}; scatter view in
  Figure~\ref{fig:cliff-scatter}.}
  \label{tab:cliff-spread}
  \centering
  \footnotesize
  \setlength{\tabcolsep}{4pt}
  \renewcommand{\arraystretch}{1.15}
  \resizebox{\textwidth}{!}{%
  \begin{tabular}{@{}l*{6}{c}@{\hskip 2.2em}r@{\hskip 1.8em}r@{}}
    \toprule
    Benchmark & Opus 4.6 & Sonnet 4.6 & Haiku 4.5 & GPT-5.4 xhigh & GPT-5.4 mini & Kimi K2.5 & Spread (pp) & SD \\
    \midrule
    SWE-Bench Verified\textsuperscript{$\dagger$}    & 79.6 & 79.2 & 73.3 & 78.2 & 73.0 & 76.8 & $\mathbf{6.6}$  & $\mathbf{2.9}$ \\
    Terminal-Bench 2.0\textsuperscript{$\dagger$}    & 65.4 & 59.1 & 41.8 & 75.1 & 60.0 & 50.8 & $\mathbf{33.3}$ & $\mathbf{11.5}$ \\
    LiveCodeBench v6\textsuperscript{$\ast$}         & 84.7 & 82.1 & 41.2 & 84.1 & 81.5 & 83.9 & $\mathbf{43.5}$ & $\mathbf{17.2}$ \\
    \midrule
    \textbf{EsoLang-Bench mean}                      & 86.9 & 66.3 & 24.7 & 99.7 & 32.5 & 11.3 & $\mathbf{88.4}$ & $\mathbf{36.0}$ \\
    \bottomrule
  \end{tabular}%
  }
\end{table}

\subsection{Strong agents discover metaprogramming strategies}
\label{sec:meta-observed}

The performance spread in Section~\ref{sec:cliff} raises a behavioral question:
what do high-performing agents do differently during the session?
Inspecting the logged trajectories shows a consistent pattern on the
low-level languages, especially Brainfuck and Befunge-98. The strongest
agents often avoid direct target-language authoring. Instead, they write
generators in a familiar host language that emit source code in the
target esolang, then run the generated programs against the local interpreter before submission. We call this metaprogramming. This subsection describes the behavior we observe in the logs; the next subsection tests whether removing it changes performance. 

This strategy is not requested by the primary prompt. The system prompts
are fixed across primary runs and reproduced in
Appendix~\ref{app:system-prompts}; the harness accepts either direct
target-language files or generated target-language files. The strategy
therefore emerges from the agent's interaction with the task substrate,
rather than from language-specific prompting.

A representative within-session switch occurs in Brainfuck E04. Opus
4.6 first submits a hand-written Brainfuck program of $1884$ bytes,
which fails the hidden tests. After the failure, it writes a Python
generator; the generated Brainfuck source is $24500$ bytes and passes
all six hidden tests. This illustrates why the strategy helps: the host
program can name and reuse structure that is implicit and fragile in raw
Brainfuck, such as cell allocation, pointer position, sign flags,
decimal-digit layout, BCD arithmetic, and conditional macros. Sample
programs and trajectory excerpts are provided in
Appendices~\ref{app:sample-programs} and~\ref{app:emergence}.

\subsection{Metaprogramming is causally important on Brainfuck and Befunge-98}
\label{sec:abl-meta}

To test whether metaprogramming merely correlates with success or
supports it, we run a no-metaprogramming variant for the two strongest
agents. In this variant, agents must author the target esolang directly
and may not use a host-language program to generate target source. All
other aspects of the protocol are held fixed.

\begin{figure}[t]
  \centering
  \includegraphics[width=\linewidth]{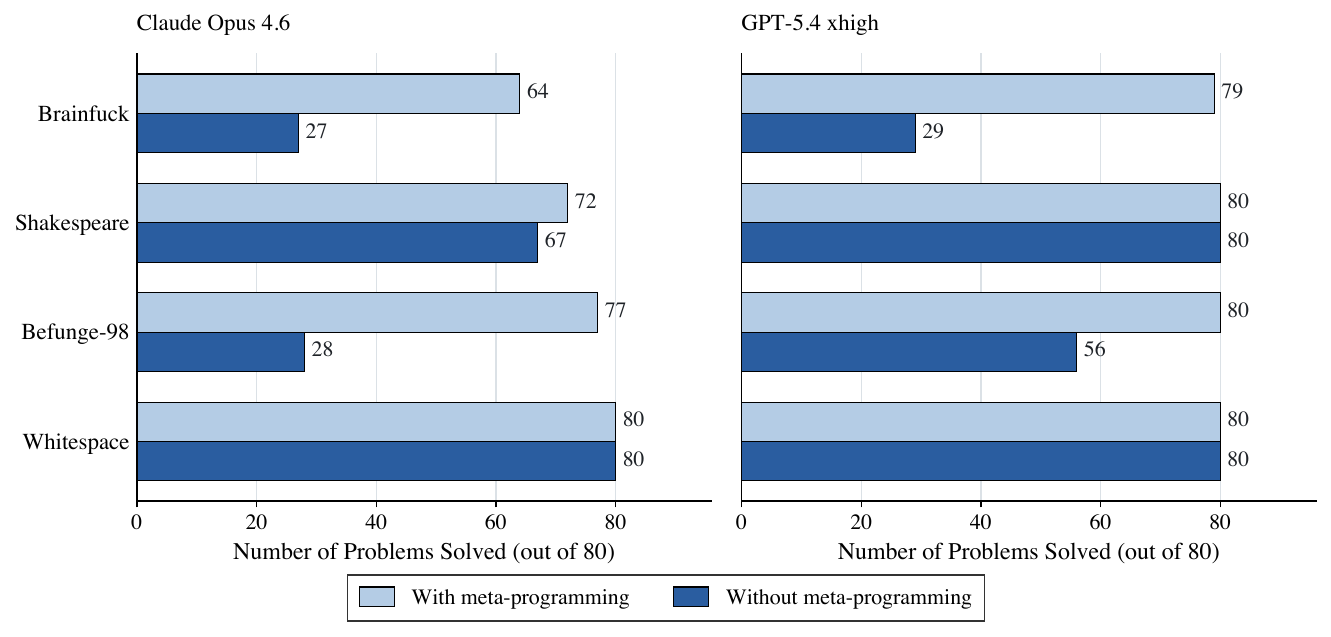}
  \caption{\textbf{Forcing direct authoring sharply reduces performance
  on Brainfuck and Befunge-98.} Solved problems out of 80 for Opus 4.6
  and GPT-5.4 xhigh with metaprogramming allowed versus forced direct
  authoring. The largest drops occur on the low-level languages where
  target programs are long and fragile.}
  \label{fig:meta-uplift}
\end{figure}

Figure~\ref{fig:meta-uplift} shows that removing metaprogramming causes
large drops on Brainfuck and Befunge-98 for both Opus 4.6 and GPT-5.4
xhigh. These are the two languages where direct authoring requires the
agent to maintain long, low-level program state across edits. In direct
Brainfuck authoring, for example, the agent must track cell offsets,
pointer position, flags, and numeric encodings implicitly while editing
raw symbols. Small local changes can invalidate this bookkeeping, and
local smoke tests often miss hidden edge cases. A host-language generator
externalizes this state into named variables and reusable functions, so
the same cell allocation, arithmetic, and branching patterns can be
emitted consistently across problems.

By contrast, Whitespace and Shakespeare are less diagnostic for this
intervention because successful solutions are often short enough or
structured enough to author directly. The result therefore supports a
narrower claim: metaprogramming is a major mechanism for high performance
on the low-level esolangs where direct target-language editing becomes
fragile. Code excerpts and trajectory examples are provided in
Appendices~\ref{app:emergence} and~\ref{app:gpt54-e04}.

\paragraph{The benefit is host-language generation, not Python specifically.}
On Brainfuck, swapping the generator host language preserves most of the
gain: Opus 4.6 solves $64$/$80$ with Python, $63$/$80$ with JavaScript, and
$55$/$80$ with Rust, while GPT-5.4 xhigh solves $79$/$80$, $77$/$80$, and $79$/$80$, respectively.
Direct authoring remains low for both agents ($27$/$80$ and $29$/$80$). Thus the
critical ingredient is access to a familiar general-purpose host
language for constructing target programs, not Python itself; the
corresponding generator code in each host language is shown in
Appendix~\ref{app:cross-language-code}.

\subsection{Strategy transfer works through executable scaffolds, but not with distilled written strategies}
\label{sec:distillation}

We next ask whether lower-performing agents fail because they lack the
high-level idea or because they cannot construct the machinery needed to
execute it. We use the strongest agents' traces (primarily Claude Opus
4.6, plus a single generic Brainfuck builder pattern from a successful
GPT-5.4 xhigh session) to create two forms of strategy transfer for
three lower-performing agents. In the \emph{text} condition, we add a
system-prompt preamble summarizing the strategy: use a generator, build
reusable primitives, verify locally, and regenerate components rather
than hand-patching target code. In the \emph{library} condition, we
additionally provide a small strategy-only host-language helper library
distilled from those traces, containing generic code-generation
primitives (cell allocator, BCD-arithmetic helpers, decimal-print
primitives, and a local Befunge-98 simulator) and a notes document; the
exact files shipped are listed in
Appendix~\ref{app:distillation-reflib-prompt}. No per-problem
generators, no solved benchmark programs, no hidden-test inputs, no
expected outputs, and no ground-truth answers are included.

Table~\ref{tab:distillation} reports solved problems out of 80 for
Brainfuck and Befunge-98. Written advice alone
produces little improvement. The reference library, in contrast,
substantially improves Sonnet 4.6 and GPT-5.4 mini, while Haiku 4.5
remains near the floor. This pattern suggests that the mid-tier agents do not mainly lack the high-level idea. They struggle to build the reusable code needed to carry it out. With that code provided, Sonnet 4.6 and GPT-5.4 mini improve sharply; Haiku 4.5 remains low.

\begin{table}[tbp]
  \caption{\textbf{Strategy transfer works through executable scaffolds, but not with distilled written strategies.}
  Results are problems solved out of 80; Base is the primary protocol.
  \emph{+Text} adds written strategy guidance distilled from Opus
  4.6's trajectory; \emph{+Lib} also provides a small strategy-only
  host-language helper library distilled from the strong-agent traces
  (primarily Opus 4.6, plus one generic GPT-5.4 xhigh Brainfuck
  builder pattern), with no per-problem generators, no solved
  programs, no hidden-test inputs, no expected outputs, and no
  ground-truth answers.}
  \label{tab:distillation}
  \centering
  \scriptsize
  \begin{tabular}{@{}lcccccccc@{}}
    \toprule
    & \multicolumn{4}{c}{Brainfuck} & \multicolumn{4}{c}{Befunge-98} \\
    \cmidrule(lr){2-5} \cmidrule(l){6-9}
    Agent & Base & +Text & +Lib & Opus & Base & +Text & +Lib & Opus \\
    \midrule
    Haiku 4.5      & 4  & 3  & 7  & 64 & 4  & 5  & 4  & 64 \\
    Sonnet 4.6     & 12 & 12 & 64 & 64 & 64 & 66 & 78 & 64 \\
    GPT-5.4 mini   & 5  & 8  & 53 & 64 & 11 & 15 & 64 & 64 \\
    \bottomrule
  \end{tabular}
\end{table}

\begin{figure}[t]
  \centering
  \includegraphics[width=0.95\linewidth]{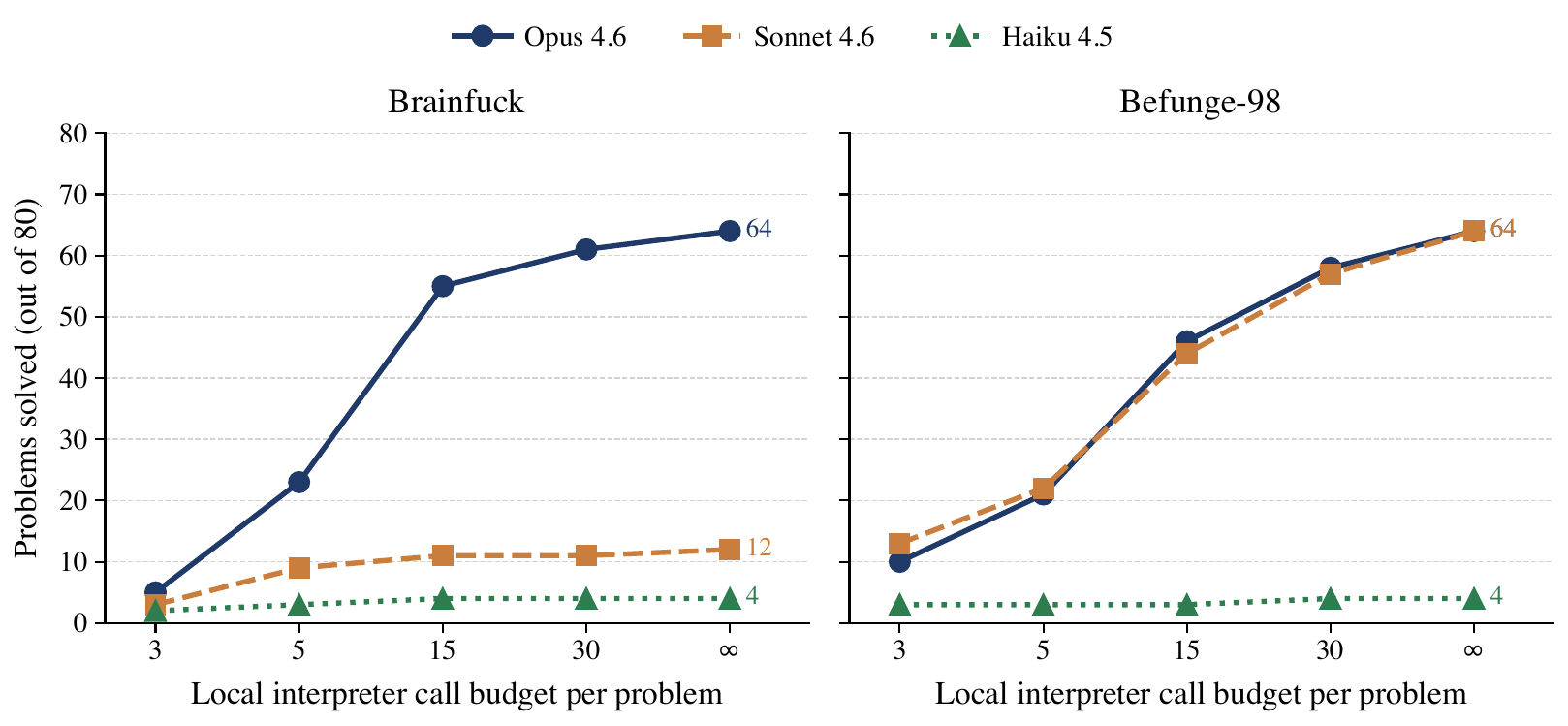}
  \caption{\textbf{More interpreter calls help only agents that can use
  feedback.} Problems solved out of 80 on Brainfuck and Befunge-98 under
  local-interpreter-call budgets of 3, 5, 15, 30, and unlimited. Opus
  improves with budget; Haiku remains near the floor; Sonnet improves on
  Befunge-98 but not Brainfuck.}
  \label{fig:scaling-curve}
\end{figure}

\subsection{Additional inference-time resources help only when agents can use them}
\label{sec:resources}

\paragraph{Interpreter-call budget.}
\label{sec:abl-budget}
We cap local interpreter calls per problem at $3$, $5$, $15$, $30$, or
unlimited, holding the task substrate, hidden submissions, and scoring
rule fixed. Figure~\ref{fig:scaling-curve} shows the result on
Brainfuck and Befunge-98. Additional interpreter access helps agents
that already convert local feedback into progress: Opus 4.6 improves on
both languages, and Sonnet 4.6 improves on Befunge-98. Agents that are
near the floor at the smallest budget remain near the floor even when
given many more local runs. Thus, tool access is not a uniform
substitute for strategy construction; it amplifies agents that can use
the feedback.

\begin{figure}[t]
  \centering
  \includegraphics[width=0.95\linewidth]{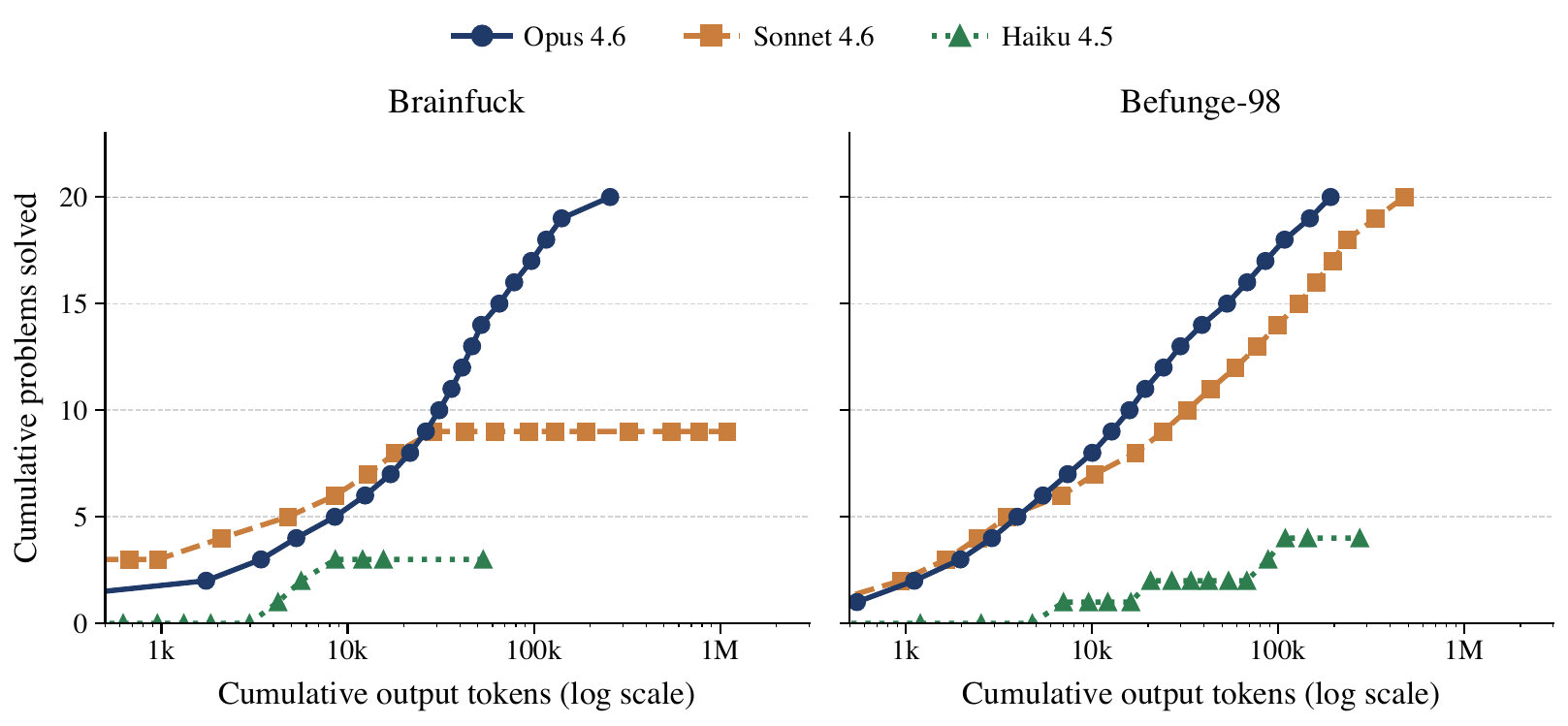}
  \caption{\textbf{Output-token use does not explain the gap.}
  Cumulative solves versus cumulative API output tokens on the first 20
  Brainfuck and Befunge-98 problems for Claude agents. Opus reaches
  20/20 on both languages with fewer tokens than Sonnet; Haiku saturates
  early.}
  \label{fig:token-curve}
\end{figure}

\paragraph{Output-token use.}
\label{sec:abl-tokens}
We also ask whether the gap is explained simply by stronger models
spending more output tokens. For the first 20 Brainfuck and Befunge-98
problems, we log cumulative API output tokens for the three Claude
agents, including extended-thinking tokens for Opus and Sonnet.
Figure~\ref{fig:token-curve} plots cumulative solves against cumulative
output tokens. Opus 4.6 solves more problems with fewer tokens than
Sonnet 4.6 on Brainfuck and reaches $20$/$20$ on Befunge-98 with roughly
half Sonnet's token use. The difference is therefore not just that Opus
spends more; it finds a reusable strategy earlier, after which
additional problems become cheaper to solve. More output budget does
not substitute for finding the strategy.

\section{Limitations and validity checks}
\label{sec:limitations}

\paragraph{Closed models and training exposure.}
The strongest agents we evaluate are closed-source, so we cannot inspect their training data,
post-training environments, or exact exposure to esolang examples. We therefore do not claim formal
distributional novelty. Our claim is operational: these are low-ecosystem programming targets
relative to mainstream languages, and they induce large differences in how deployed coding agents
adapt during a session. Appendix~\ref{app:contamination} reports public-code prevalence and
$n$-gram overlap analyses showing orders-of-magnitude less public-code
presence for these esolangs than for mainstream languages.
Hidden tests, isolated workspaces, and fixed forward sessions reduce shallow
memorization but do not prove zero exposure.

\paragraph{Artificial but controlled proxies for testing adaptation.}
The esolangs are artificial, but that is what makes them useful here. They are public, fully specified,
runnable, automatically graded languages with unusual syntax, execution rules, and debugging
surfaces. This gives us a controlled proxy for a practical pressure that is otherwise hard to study
publicly: whether an agent can build, test, and revise a working interface model when familiar
language and library priors are weak. We therefore treat success on these tasks as evidence about
adaptation to low-ecosystem executable interfaces, not as evidence that esolangs themselves are
important production targets.

\paragraph{Harnesses and mechanism scope.}
We evaluate deployed coding agents rather than bare language models, because tool use, file editing,
shell access, and workspace management are part of the systems users actually run. The tradeoff is
that Claude Code, Codex, and OpenCode are not identical internally. Our protocol fixes the
benchmark-facing interface: every agent receives the same problem sequence, local interpreters,
hidden-test rule, and scoring criterion. Selected OpenCode re-runs preserve the qualitative ordering
(Appendix~\ref{app:cross-harness}), making the separation unlikely to be wrapper-only.
The metaprogramming claim is specific to Brainfuck and Befunge-98 where direct authoring is
fragile, not Whitespace or Shakespeare.

\section{Related work}
\label{sec:related}

\paragraph{Code and agentic coding benchmarks.}
Execution-based grading is central to modern code evaluation, from
HumanEval~\citep{chen2021codex}, MBPP~\citep{austin2021program},
APPS~\citep{hendrycks2021apps}, DS-1000~\citep{lai2023ds1000}, and
MultiPL-E~\citep{cassano2023multiple} to LiveCodeBench~\citep{jain2024livecodebench},
BigCodeBench~\citep{zhuo2024bigcodebench}, OJBench~\citep{ojbench2025},
SciCode~\citep{tian2024scicode}, MHPP~\citep{dai2024mhpp}, and
SWE-bench~\citep{jimenez2024swebench}. Agentic benchmarks extend this to multi-step work in repositories, terminals, desktops, browsers,
and research environments~\citep{yang2024sweagent,xia2024agentless,zhang2024autocoderover,
merrill2026terminalbench,xie2024osworld,drouin2024workarena,koh2024visualwebarena,
he2024webvoyager,starace2025paperbench,chan2024mlebench}.
These settings are realistic but mix many factors: repository navigation, dependency management,
long-horizon planning, environment quirks, and familiarity with public software ecosystems. We
inherit executable grading, but change the controlled variable: using EsoLang-Bench
~\citep{esolangbench2026}, we keep the tasks simple and vary the familiarity of the executable
interface itself. ARC-AGI-3~\citep{arcagi32026} is close in spirit, but stresses rule inference in
unfamiliar games; we stress unfamiliar programming interfaces with explicit task specifications.

\paragraph{Tools, feedback, and language transfer.}
Prior work shows that tools, feedback, and intermediate computation can improve LLM performance
~\citep{yao2022react,schick2023toolformer,shinn2023reflexion,madaan2023selfrefine,
yao2023tree,gao2023pal,chen2022program,chen2023selfdebug,nye2021show}, and
benchmarks such as TAU-bench~\citep{yao2025taubench} and BFCL~\citep{patil2025bfcl}
evaluate interface following and tool use. Our question is which agents can turn local execution and
workspace persistence into reliable program construction when the target language is unfamiliar.
The cross-host experiment also connects to multilingual code generation and translation
~\citep{cassano2023multiple,paul2024ircoder,twist2025llmslovepython,
roziere2020transcoder,ahmad2021plbart,wang2021codet5,fried2022incoder,
nijkamp2022codegen,li2023starcoder,roziere2023codellama}. Our strongest agents often do
not simply generate the target language directly; they write generators in a familiar host language
and treat the target language as generated output.

\paragraph{Benchmark validity and long-tail coding.}
Benchmark-design and contamination work
~\citep{ribeiro2020checklist,nie2020anli,ye2021crossfit,bowman2021benchmarking,
oren2024proving,deng2024contamination,xu2024survey} warns that high scores can mask
weaknesses and that training overlap is hard to rule out for proprietary models. We therefore avoid
formal OOD claims and report public-code frequency and $n$-gram overlap analyses. The motivation
also connects to long-tail production coding, where agents face internal DSLs, generated APIs,
proprietary configuration formats, and platform-specific frameworks that are sparse in public training
data. Industry evaluations such as Convex Evals~\citep{convexevals} and
Fullstack-Bench~\citep{fullstackbench} document failures on such platform-specific invariants. Our
esolang setting makes an analogous pressure public, runnable, inspectable, and automatically graded.

\section{Conclusion}
\label{sec:discussion}

Unfamiliar programming languages make a normally hidden agent capability visible. When the
target language is low-ecosystem, success requires more than writing code in a familiar software
environment. The agent has to work out how the language behaves, test candidate programs, revise
failures, and decide when a solution is ready. Because the problems are short standard tasks and
every agent receives the same benchmark-facing interface, the separation reflects how well agents
turn tools, feedback, and workspace state into a working strategy for the target language.

For the strongest agents, the clearest strategy is metaprogramming: host-language generators,
reusable primitives, and local verification loops. The no-metaprogramming ablation shows this
machinery is causally important on Brainfuck and Befunge-98, where direct authoring is long and
fragile. Strategy transfer sharpens the point: written advice does little, while working helper code
lifts mid-tier agents and still fails on the weakest. The key capability is not knowing that a strategy
should help, but building and debugging machinery that works under the target language's rules.

Agentic generalization in this setting means reorganizing an unfamiliar problem into a form the
agent can solve. The strongest agents do not only retrieve familiar patterns; they create intermediate
code, tests, and reusable structure that make the target language usable. Real deployments often
involve internal DSLs, generated APIs, proprietary configuration formats, and local tool conventions
that are sparse in public code. Making this capability reliable in smaller or open-source agents should
become a target for training, distillation, and model analysis.

\bibliographystyle{plainnat}
\bibliography{references}

@article{chen2021codex,
  title={Evaluating Large Language Models Trained on Code},
  author={Chen, Mark and Tworek, Jerry and Jun, Heewoo and Yuan, Qiming and Pinto, Henrique Ponde de Oliveira and Kaplan, Jared and Edwards, Harri and Burda, Yuri and Joseph, Nicholas and Brockman, Greg and Ray, Alex and Puri, Raul and Krueger, Gretchen and Petrov, Michael and Khlaaf, Heidy and Sastry, Girish and Mishkin, Pamela and Chan, Brooke and Gray, Scott and Ryder, Nick and Pavlov, Mikhail and Power, Alethea and Kaiser, Lukasz and Bavarian, Mohammad and Winter, Clemens and Tillet, Philippe and Such, Felipe Petroski and Cummings, Dave and Plappert, Matthias and Chantzis, Fotios and Barnes, Elizabeth and Herbert-Voss, Ariel and Guss, William Hebgen and Nichol, Alex and Paino, Alex and Tezak, Nikolas and Tang, Jie and Babuschkin, Igor and Balaji, Suchir and Jain, Shantanu and Saunders, William and Hesse, Christopher and Carr, Andrew N. and Leike, Jan and Achiam, Josh and Misra, Vedant and Morikawa, Evan and Radford, Alec and Knight, Matthew and Brundage, Miles and Murati, Mira and Mayer, Katie and Welinder, Peter and McGrew, Bob and Amodei, Dario and McCandlish, Sam and Sutskever, Ilya and Zaremba, Wojciech},
  journal={arXiv preprint arXiv:2107.03374},
  year={2021}
}

@article{austin2021program,
  title={Program Synthesis with Large Language Models},
  author={Austin, Jacob and Odena, Augustus and Nye, Maxwell and Bosma, Maarten and Michalewski, Henryk and Dohan, David and Jiang, Ellen and Cai, Carrie and Terry, Michael and Le, Quoc and Sutton, Charles},
  journal={arXiv preprint arXiv:2108.07732},
  year={2021}
}

@inproceedings{hendrycks2021apps,
  title={Measuring Coding Challenge Competence With {APPS}},
  author={Hendrycks, Dan and Basart, Steven and Kadavath, Saurav and Mazeika, Mantas and Arora, Akul and Guo, Ethan and Burns, Collin and Puranik, Samir and He, Horace and Song, Dawn and Steinhardt, Jacob},
  booktitle={Advances in Neural Information Processing Systems Datasets and Benchmarks Track},
  year={2021}
}

@inproceedings{lai2023ds1000,
  title={{DS-1000}: A Natural and Reliable Benchmark for Data Science Code Generation},
  author={Lai, Yuhang and Li, Chengxi and Wang, Yiming and Zhang, Tianyi and Zhong, Ruiqi and Zettlemoyer, Luke and Yih, Wen-Tau and Fried, Daniel and Wang, Sida and Yu, Tao},
  booktitle={Proceedings of the 40th International Conference on Machine Learning},
  series={Proceedings of Machine Learning Research},
  volume={202},
  pages={18319--18345},
  year={2023}
}

@article{cassano2023multiple,
  title={{MultiPL-E}: A Scalable and Polyglot Approach to Benchmarking Neural Code Generation},
  author={Cassano, Federico and Gouwar, John and Nguyen, Daniel and Nguyen, Sydney and Phipps-Costin, Luna and Pinckney, Donald and Yee, Ming-Ho and Zi, Yangtian and Anderson, Carolyn Jane and Feldman, Molly Q. and Guha, Arjun and Greenberg, Michael and Jangda, Abhinav},
  journal={IEEE Transactions on Software Engineering},
  volume={49},
  number={7},
  pages={3675--3691},
  year={2023},
  doi={10.1109/TSE.2023.3267446}
}

@inproceedings{jain2024livecodebench,
  title={{LiveCodeBench}: Holistic and Contamination Free Evaluation of Large Language Models for Code},
  author={Jain, Naman and Han, King and Gu, Alex and Li, Wen-Ding and Yan, Fanjia and Zhang, Tianjun and Wang, Sida and Solar-Lezama, Armando and Sen, Koushik and Stoica, Ion},
  booktitle={International Conference on Learning Representations},
  year={2025},
  url={https://openreview.net/forum?id=chfJJYC3iL}
}

@inproceedings{zhuo2024bigcodebench,
  title={{BigCodeBench}: Benchmarking Code Generation with Diverse Function Calls and Complex Instructions},
  author={Zhuo, Terry Yue and Vu, Minh Chien and Chim, Jenny and Hu, Han and Yu, Wenhao and Widyasari, Ratnadira and Yusuf, Imam Nur Bani and Zhan, Haolan and He, Junda and Paul, Indraneil and Brunner, Simon and Gong, Chen and Hoang, James and Zebaze, Armel Randy and Hong, Xiaoheng and Li, Wen-Ding and Kaddour, Jean and Xu, Ming and Zhang, Zhihan and Yadav, Prateek and Jain, Naman and Gu, Alex and Cheng, Zhoujun and Liu, Jiawei and Liu, Qian and Wang, Zijian and Hui, Binyuan and Muennighoff, Niklas and Lo, David and Fried, Daniel and Du, Xiaoning and de Vries, Harm and von Werra, Leandro},
  booktitle={International Conference on Learning Representations},
  year={2025},
  url={https://openreview.net/forum?id=YrycTjllL0}
}

@article{tian2024scicode,
  title={{SciCode}: A Research Coding Benchmark Curated by Scientists},
  author={Tian, Minyang and Gao, Luyu and Zhang, Shizhuo Dylan and Chen, Xinan and Fan, Cunwei and Guo, Xuefei and Haas, Roland and Ji, Pan and Krongchon, Kittithat and Li, Yao and Liu, Shengyan and Luo, Di and Ma, Yutao and Tong, Hao and Trinh, Kha and Tian, Chenyu and Wang, Zihan and Wu, Bohao and Xiong, Yanyu and Yin, Shengzhu and Zhu, Minhui and Lieret, Kilian and Lu, Yanxin and Liu, Genglin and Du, Yufeng and Tao, Tianhua and Press, Ofir and Callan, Jamie and Huerta, Eliu and Peng, Hao},
  journal={arXiv preprint arXiv:2407.13168},
  year={2024}
}

@article{ojbench2025,
  title={{OJBench}: A Competition Level Code Benchmark For Large Language Models},
  author={Wang, Zhexu and Liu, Yiping and Wang, Yejie and He, Wenyang and Gao, Bofei and Diao, Muxi and Chen, Yanxu and Fu, Kelin and Sung, Flood and Yang, Zhilin and Liu, Tianyu and Xu, Weiran},
  journal={arXiv preprint arXiv:2506.16395},
  year={2025}
}

@inproceedings{jimenez2024swebench,
  title={{SWE-bench}: Can Language Models Resolve Real-World {GitHub} Issues?},
  author={Jimenez, Carlos E. and Yang, John and Wettig, Alexander and Yao, Shunyu and Pei, Kexin and Press, Ofir and Narasimhan, Karthik},
  booktitle={International Conference on Learning Representations},
  year={2024}
}

@article{merrill2026terminalbench,
  title={{Terminal-Bench}: Benchmarking Agents on Hard, Realistic Tasks in Command Line Interfaces},
  author={Merrill, Mike A. and Shaw, Alexander G. and Carlini, Nicholas and Li, Boxuan and Raj, Harsh and Bercovich, Ivan and Shi, Lin and Shin, Jeong Yeon and Walshe, Thomas and Buchanan, E. Kelly and Shen, Junhong and Ye, Guanghao and Lin, Haowei and Poulos, Jason and Wang, Maoyu and Nezhurina, Marianna and Jitsev, Jenia and Lu, Di and Mastromichalakis, Orfeas Menis and Xu, Zhiwei and Chen, Zizhao and Liu, Yue and Zhang, Robert and Chen, Leon Liangyu and Kashyap, Anurag and Uslu, Jan-Lucas and Li, Jeffrey and Wu, Jianbo and Yan, Minghao and Bian, Song and others},
  journal={arXiv preprint arXiv:2601.11868},
  year={2026},
  url={https://arxiv.org/abs/2601.11868}
}

@inproceedings{starace2025paperbench,
  title={{PaperBench}: Evaluating {AI}'s Ability to Replicate {AI} Research},
  author={Starace, Giulio and Jaffe, Oliver and Sherburn, Dane and Aung, James and Chan, Jun Shern and Maksin, Leon and Dias, Rachel and Mays, Evan and Kinsella, Benjamin and Thompson, Wyatt and Heidecke, Johannes and Glaese, Amelia and Patwardhan, Tejal},
  booktitle={Proceedings of the 42nd International Conference on Machine Learning},
  series={Proceedings of Machine Learning Research},
  volume={267},
  pages={56843--56873},
  year={2025},
  url={https://proceedings.mlr.press/v267/starace25a.html}
}

@inproceedings{chan2024mlebench,
  title={{MLE-bench}: Evaluating Machine Learning Agents on Machine Learning Engineering},
  author={Chan, Jun Shern and Chowdhury, Neil and Jaffe, Oliver and Aung, James and Sherburn, Dane and Mays, Evan and Starace, Giulio and Liu, Kevin and Maksin, Leon and Patwardhan, Tejal and Madry, Aleksander and Weng, Lilian},
  booktitle={International Conference on Learning Representations},
  year={2025},
  url={https://openreview.net/forum?id=6s5uXNWGIh}
}

@inproceedings{xie2024osworld,
  title={{OSWorld}: Benchmarking Multimodal Agents for Open-Ended Tasks in Real Computer Environments},
  author={Xie, Tianbao and Zhang, Danyang and Chen, Jixuan and Li, Xiaochuan and Zhao, Siheng and Cao, Ruisheng and Hua, Toh Jing and Cheng, Zhoujun and Shin, Dongchan and Lei, Fangyu and Liu, Yitao and Xu, Yiheng and Zhou, Shuyan and Savarese, Silvio and Xiong, Caiming and Zhong, Victor and Yu, Tao},
  booktitle={Advances in Neural Information Processing Systems},
  volume={37},
  year={2024},
  note={Datasets and Benchmarks Track},
  url={https://openreview.net/forum?id=tN61DTr4Ed}
}

@inproceedings{yang2024sweagent,
  title={{SWE-agent}: Agent-Computer Interfaces Enable Automated Software Engineering},
  author={Yang, John and Jimenez, Carlos E. and Wettig, Alexander and Lieret, Kilian and Yao, Shunyu and Narasimhan, Karthik and Press, Ofir},
  booktitle={Advances in Neural Information Processing Systems},
  volume={37},
  year={2024}
}

@article{xia2024agentless,
  title={{Agentless}: Demystifying {LLM}-based Software Engineering Agents},
  author={Xia, Chunqiu Steven and Deng, Yinlin and Dunn, Soren and Zhang, Lingming},
  journal={arXiv preprint arXiv:2407.01489},
  year={2024}
}

@inproceedings{zhang2024autocoderover,
  title={{AutoCodeRover}: Autonomous Program Improvement},
  author={Zhang, Yuntong and Ruan, Haifeng and Fan, Zhiyu and Roychoudhury, Abhik},
  booktitle={Proceedings of the 33rd ACM SIGSOFT International Symposium on Software Testing and Analysis},
  year={2024},
  doi={10.1145/3650212.3680384}
}

@inproceedings{yao2025taubench,
  title={$\tau$-bench: A Benchmark for Tool-Agent-User Interaction in Real-World Domains},
  author={Yao, Shunyu and Shinn, Noah and Razavi, Pedram and Narasimhan, Karthik R.},
  booktitle={International Conference on Learning Representations},
  year={2025},
  url={https://openreview.net/forum?id=roNSXZpUDN}
}

@inproceedings{patil2025bfcl,
  title={The Berkeley Function Calling Leaderboard ({BFCL}): From Tool Use to Agentic Evaluation of Large Language Models},
  author={Patil, Shishir G. and Mao, Huanzhi and Yan, Fanjia and Ji, Charlie Cheng-Jie and Suresh, Vishnu and Stoica, Ion and Gonzalez, Joseph E.},
  booktitle={Proceedings of the 42nd International Conference on Machine Learning},
  series={Proceedings of Machine Learning Research},
  volume={267},
  pages={48371--48392},
  year={2025},
  url={https://proceedings.mlr.press/v267/patil25a.html}
}

@inproceedings{drouin2024workarena,
  title={{WorkArena}: How Capable are Web Agents at Solving Common Knowledge Work Tasks?},
  author={Drouin, Alexandre and Gasse, Maxime and Caccia, Massimo and Laradji, Issam H. and Del Verme, Manuel and Marty, Tom and Vazquez, David and Chapados, Nicolas and Lacoste, Alexandre},
  booktitle={Proceedings of the 41st International Conference on Machine Learning},
  series={Proceedings of Machine Learning Research},
  volume={235},
  pages={11642--11662},
  year={2024},
  url={https://proceedings.mlr.press/v235/drouin24a.html}
}

@inproceedings{koh2024visualwebarena,
  title={{VisualWebArena}: Evaluating Multimodal Agents on Realistic Visual Web Tasks},
  author={Koh, Jing Yu and Lo, Robert and Jang, Lawrence and Duvvur, Vikram and Lim, Ming and Huang, Po-Yu and Neubig, Graham and Zhou, Shuyan and Salakhutdinov, Russ and Fried, Daniel},
  booktitle={Proceedings of the 62nd Annual Meeting of the Association for Computational Linguistics (Volume 1: Long Papers)},
  pages={881--905},
  year={2024},
  doi={10.18653/v1/2024.acl-long.50}
}

@inproceedings{he2024webvoyager,
  title={{WebVoyager}: Building an End-to-End Web Agent with Large Multimodal Models},
  author={He, Hongliang and Yao, Wenlin and Ma, Kaixin and Yu, Wenhao and Dai, Yong and Zhang, Hongming and Lan, Zhenzhong and Yu, Dong},
  booktitle={Proceedings of the 62nd Annual Meeting of the Association for Computational Linguistics (Volume 1: Long Papers)},
  pages={6864--6890},
  year={2024},
  doi={10.18653/v1/2024.acl-long.371}
}

@inproceedings{yao2022react,
  title={{ReAct}: Synergizing Reasoning and Acting in Language Models},
  author={Yao, Shunyu and Zhao, Jeffrey and Yu, Dian and Du, Nan and Shafran, Izhak and Narasimhan, Karthik and Cao, Yuan},
  booktitle={International Conference on Learning Representations},
  year={2023}
}

@inproceedings{schick2023toolformer,
  title={{Toolformer}: Language Models Can Teach Themselves to Use Tools},
  author={Schick, Timo and Dwivedi-Yu, Jane and Dess{\i}, Roberto and Raileanu, Roberta and Lomeli, Maria and Hambro, Eric and Zettlemoyer, Luke and Cancedda, Nicola and Scialom, Thomas},
  booktitle={Advances in Neural Information Processing Systems},
  volume={36},
  year={2023}
}

@inproceedings{shinn2023reflexion,
  title={{Reflexion}: Language Agents with Verbal Reinforcement Learning},
  author={Shinn, Noah and Cassano, Federico and Berman, Edward and Gopinath, Ashwin and Narasimhan, Karthik and Yao, Shunyu},
  booktitle={Advances in Neural Information Processing Systems},
  volume={36},
  year={2023}
}

@inproceedings{madaan2023selfrefine,
  title={{Self-Refine}: Iterative Refinement with Self-Feedback},
  author={Madaan, Aman and Tandon, Niket and Gupta, Prakhar and Hallinan, Skyler and Gao, Luyu and Wiegreffe, Sarah and Alon, Uri and Dziri, Nouha and Prabhumoye, Shrimai and Yang, Yiming and Gupta, Shashank and Majumder, Bodhisattwa Prasad and Hermann, Katherine and Welleck, Sean and Yazdanbakhsh, Amir and Clark, Peter},
  booktitle={Advances in Neural Information Processing Systems},
  volume={36},
  year={2023}
}

@inproceedings{yao2023tree,
  title={Tree of Thoughts: Deliberate Problem Solving with Large Language Models},
  author={Yao, Shunyu and Yu, Dian and Zhao, Jeffrey and Shafran, Izhak and Griffiths, Thomas L. and Cao, Yuan and Narasimhan, Karthik},
  booktitle={Advances in Neural Information Processing Systems},
  volume={36},
  year={2023}
}

@inproceedings{gao2023pal,
  title={{PAL}: Program-aided Language Models},
  author={Gao, Luyu and Madaan, Aman and Zhou, Shuyan and Alon, Uri and Liu, Pengfei and Yang, Yiming and Callan, Jamie and Neubig, Graham},
  booktitle={Proceedings of the 40th International Conference on Machine Learning},
  series={Proceedings of Machine Learning Research},
  volume={202},
  pages={10764--10799},
  year={2023}
}

@article{chen2022program,
  title={Program of Thoughts Prompting: Disentangling Computation from Reasoning for Numerical Reasoning Tasks},
  author={Chen, Wenhu and Ma, Xueguang and Wang, Xinyi and Cohen, William W.},
  journal={Transactions on Machine Learning Research},
  year={2023},
  note={arXiv:2211.12588}
}

@inproceedings{chen2023selfdebug,
  title={Teaching Large Language Models to Self-Debug},
  author={Chen, Xinyun and Lin, Maxwell and Sch{\"a}rli, Nathanael and Zhou, Denny},
  booktitle={International Conference on Learning Representations},
  year={2024}
}

@article{nye2021show,
  title={Show Your Work: Scratchpads for Intermediate Computation with Language Models},
  author={Nye, Maxwell and Andreassen, Anders Johan and Gur-Ari, Guy and Michalewski, Henryk and Austin, Jacob and Bieber, David and Dohan, David and Lewkowycz, Aitor and Bosma, Maarten and Luan, David and Sutton, Charles and Odena, Augustus},
  journal={arXiv preprint arXiv:2112.00114},
  year={2021}
}

@inproceedings{ribeiro2020checklist,
  title={Beyond Accuracy: Behavioral Testing of {NLP} Models with {CheckList}},
  author={Ribeiro, Marco Tulio and Wu, Tongshuang and Guestrin, Carlos and Singh, Sameer},
  booktitle={Proceedings of the 58th Annual Meeting of the Association for Computational Linguistics},
  pages={4902--4912},
  year={2020},
  doi={10.18653/v1/2020.acl-main.442}
}

@inproceedings{nie2020anli,
  title={Adversarial {NLI}: A New Benchmark for Natural Language Understanding},
  author={Nie, Yixin and Williams, Adina and Dinan, Emily and Bansal, Mohit and Weston, Jason and Kiela, Douwe},
  booktitle={Proceedings of the 58th Annual Meeting of the Association for Computational Linguistics},
  pages={4885--4901},
  year={2020},
  doi={10.18653/v1/2020.acl-main.441}
}

@inproceedings{ye2021crossfit,
  title={{CrossFit}: A Few-shot Learning Challenge for Cross-task Generalization in {NLP}},
  author={Ye, Qinyuan and Lin, Bill Yuchen and Ren, Xiang},
  booktitle={Proceedings of the 2021 Conference on Empirical Methods in Natural Language Processing},
  pages={7163--7189},
  year={2021},
  doi={10.18653/v1/2021.emnlp-main.572}
}

@inproceedings{bowman2021benchmarking,
  title={What Will it Take to Fix Benchmarking in Natural Language Understanding?},
  author={Bowman, Samuel R. and Dahl, George E.},
  booktitle={Proceedings of the 2021 Conference of the North American Chapter of the Association for Computational Linguistics: Human Language Technologies},
  pages={4843--4855},
  year={2021},
  doi={10.18653/v1/2021.naacl-main.385}
}

@inproceedings{oren2024proving,
  title={Proving Test Set Contamination in Black-Box Language Models},
  author={Oren, Yonatan and Meister, Nicole and Chatterji, Niladri S. and Ladhak, Faisal and Hashimoto, Tatsunori B.},
  booktitle={International Conference on Learning Representations},
  year={2024}
}

@inproceedings{deng2024contamination,
  title={Investigating Data Contamination in Modern Benchmarks for Large Language Models},
  author={Deng, Chunyuan and Zhao, Yilun and Tang, Xiangru and Gerstein, Mark and Cohan, Arman},
  booktitle={Proceedings of the 2024 Conference of the North American Chapter of the Association for Computational Linguistics: Human Language Technologies (Volume 1: Long Papers)},
  pages={8706--8719},
  year={2024},
  doi={10.18653/v1/2024.naacl-long.482}
}

@article{xu2024survey,
  title={Benchmark Data Contamination of Large Language Models: A Survey},
  author={Xu, Cheng and Guan, Shuhao and Greene, Derek and Kechadi, M. Tahar},
  journal={arXiv preprint arXiv:2406.04244},
  year={2024}
}

@inproceedings{paul2024ircoder,
  title={{IRCoder}: Intermediate Representations Make Language Models Robust Multilingual Code Generators},
  author={Paul, Indraneil and Glava{\v{s}}, Goran and Gurevych, Iryna},
  booktitle={Proceedings of the 62nd Annual Meeting of the Association for Computational Linguistics (Volume 1: Long Papers)},
  pages={15023--15041},
  year={2024},
  doi={10.18653/v1/2024.acl-long.802}
}

@article{twist2025llmslovepython,
  title={A Study of LLMs' Preferences for Libraries and Programming Languages},
  author={Twist, Lukas and Zhang, Jie M. and Harman, Mark and Syme, Don and Noppen, Joost and Yannakoudakis, Helen and Nauck, Detlef},
  journal={arXiv preprint arXiv:2503.17181},
  year={2025},
  note={To appear in Findings of ACL 2026}
}

@inproceedings{roziere2020transcoder,
  title={Unsupervised Translation of Programming Languages},
  author={Roziere, Baptiste and Lachaux, Marie-Anne and Chanussot, Lowik and Lample, Guillaume},
  booktitle={Advances in Neural Information Processing Systems},
  volume={33},
  year={2020}
}

@inproceedings{ahmad2021plbart,
  title={Unified Pre-training for Program Understanding and Generation},
  author={Ahmad, Wasi Uddin and Chakraborty, Saikat and Ray, Baishakhi and Chang, Kai-Wei},
  booktitle={Proceedings of the 2021 Conference of the North American Chapter of the Association for Computational Linguistics: Human Language Technologies},
  pages={2655--2668},
  year={2021},
  doi={10.18653/v1/2021.naacl-main.211}
}

@inproceedings{wang2021codet5,
  title={{CodeT5}: Identifier-aware Unified Pre-trained Encoder-Decoder Models for Code Understanding and Generation},
  author={Wang, Yue and Wang, Weishi and Joty, Shafiq and Hoi, Steven C. H.},
  booktitle={Proceedings of the 2021 Conference on Empirical Methods in Natural Language Processing},
  pages={8696--8708},
  year={2021},
  doi={10.18653/v1/2021.emnlp-main.685}
}

@inproceedings{fried2022incoder,
  title={{InCoder}: A Generative Model for Code Infilling and Synthesis},
  author={Fried, Daniel and Aghajanyan, Armen and Lin, Jessy and Wang, Sida and Wallace, Eric and Shi, Freda and Zhong, Ruiqi and Yih, Wen-tau and Zettlemoyer, Luke and Lewis, Mike},
  booktitle={International Conference on Learning Representations},
  year={2023},
  note={arXiv:2204.05999}
}

@article{nijkamp2022codegen,
  title={{CodeGen}: An Open Large Language Model for Code with Multi-Turn Program Synthesis},
  author={Nijkamp, Erik and Pang, Bo and Hayashi, Hiroaki and Tu, Lifu and Wang, Huan and Zhou, Yingbo and Savarese, Silvio and Xiong, Caiming},
  journal={arXiv preprint arXiv:2203.13474},
  year={2022}
}

@article{li2023starcoder,
  title={{StarCoder}: may the source be with you!},
  author={Li, Raymond and Allal, Loubna Ben and Zi, Yangtian and Muennighoff, Niklas and Kocetkov, Denis and Mou, Chenghao and Marone, Marc and Akiki, Christopher and Li, Jia and Chim, Jenny and Liu, Qian and Zheltonozhskii, Evgenii and Zhuo, Terry Yue and Wang, Thomas and Dehaene, Olivier and Davaadorj, Mergen and Lamy-Poirier, Joel and Monteiro, Joao and Shliazhko, Oleh and Gontier, Nicolas and Meade, Nicholas and Zebaze, Armel Randy and Yee, Ming-Ho and Umapathi, Logesh Kumar and Zhu, Jian and Lipkin, Ben and Oblokulov, Muhtasham and Wang, Zhiruo and others},
  journal={arXiv preprint arXiv:2305.06161},
  year={2023}
}

@article{roziere2023codellama,
  title={Code {Llama}: Open Foundation Models for Code},
  author={Roziere, Baptiste and Gehring, Jonas and Gloeckle, Fabian and Sootla, Sten and Gat, Itai and Tan, Xiaoqing Ellen and Adi, Yossi and Liu, Jingyu and Sauvestre, Romain and Remez, Tal and Rapin, Jeremy and Kozhevnikov, Artyom and Evtimov, Ivan and Bitton, Joanna and Bhatt, Manish and Ferrer, Cristian Canton and Grattafiori, Aaron and Xiong, Wenhan and Defossez, Alexandre and Copet, Jade and Azhar, Faisal and Touvron, Hugo and Martin, Louis and Usunier, Nicolas and Scialom, Thomas and Synnaeve, Gabriel},
  journal={arXiv preprint arXiv:2308.12950},
  year={2023}
}

@article{esolangbench2026,
  title={{EsoLang-Bench}: Evaluating Genuine Reasoning in Large Language Models via Esoteric Programming Languages},
  author={Sharma, Aman and Chopra, Paras},
  journal={arXiv preprint arXiv:2603.09678},
  year={2026}
}

@misc{esolangbench_hf_2026,
  title={{Lossfunk/Esolang-Bench}: Hugging Face Dataset},
  author={{Lossfunk}},
  year={2026},
  howpublished={Hugging Face Datasets},
  url={https://huggingface.co/datasets/Lossfunk/Esolang-Bench},
  note={CC BY 4.0. Accessed 2026-05-07.}
}

@misc{arcagi32026,
  title={{ARC-AGI-3}: A New Challenge for Frontier Agentic Intelligence},
  author={{ARC Prize Foundation}},
  year={2026},
  eprint={2603.24621},
  archivePrefix={arXiv},
  primaryClass={cs.AI},
  url={https://arxiv.org/abs/2603.24621}
}

@misc{convexevals,
  title={{Convex Evals}: Behind the Scenes of {AI} Coding with {Convex}},
  author={Hunt, Jordan},
  year={2025},
  howpublished={Convex Stack Blog},
  url={https://stack.convex.dev/convex-evals},
  note={Accessed 2026-04-22}
}

@misc{fullstackbench,
  title={Introducing {Fullstack-Bench}},
  author={Jayakar, Sujay},
  year={2025},
  howpublished={Convex Stack Blog},
  url={https://stack.convex.dev/introducing-fullstack-bench},
  note={Accessed 2026-04-22}
}

@article{dai2024mhpp,
  title={{MHPP}: Exploring the Capabilities and Limitations of Language Models Beyond Basic Code Generation},
  author={Dai, Jianbo and Lu, Jianqiao and Feng, Yunlong and Zeng, Guangtao and Ruan, Rongju and Cheng, Ming and Huang, Dong and Tan, Haochen and Guo, Zhijiang},
  journal={arXiv preprint arXiv:2405.11430},
  year={2024}
}

@inproceedings{du2024classeval,
  title={Evaluating Large Language Models in Class-Level Code Generation},
  author={Du, Xueying and Liu, Mingwei and Wang, Kaixin and Wang, Hanlin and Liu, Junwei and Chen, Yixuan and Feng, Jiayi and Sha, Chaofeng and Peng, Xin and Lou, Yiling},
  booktitle={Proceedings of the 46th IEEE/ACM International Conference on Software Engineering (ICSE)},
  year={2024}
}

@inproceedings{yu2024codereval,
  title={{CoderEval}: A Benchmark of Pragmatic Code Generation with Generative Pre-Trained Models},
  author={Yu, Hao and Shen, Bo and Ran, Dezhi and Zhang, Jiaxin and Zhang, Qi and Ma, Yuchi and Liang, Guangtai and Li, Ying and Wang, Qianxiang and Xie, Tao},
  booktitle={Proceedings of the 46th IEEE/ACM International Conference on Software Engineering (ICSE)},
  year={2024},
  note={arXiv:2302.00288}
}

@article{motwani2026longcot,
  title={{LongCoT}: Benchmarking Long-Horizon Chain-of-Thought Reasoning},
  author={Motwani, Sumeet Ramesh and Nichols, Daniel and London, Charles and Li, Peggy and Pizzati, Fabio and Blake, Acer and Hammoud, Hasan and McDonald, Tavish and Naik, Akshat and Ivanova, Alesia and Baskaran, Vignesh and Laptev, Ivan and Glatt, Ruben and Ben-Nun, Tal and Torr, Philip and Jaques, Natasha and Prabhu, Ameya and Bartoldson, Brian and Kailkhura, Bhavya and Schroeder de Witt, Christian},
  journal={arXiv preprint arXiv:2604.14140},
  year={2026},
  url={https://arxiv.org/abs/2604.14140}
}

@misc{anthropic_models_2026,
  title={{Claude Opus 4.6}, {Claude Sonnet 4.6}, and {Claude Haiku 4.5}: Model Overview},
  author={{Anthropic}},
  year={2026},
  howpublished={Anthropic model overview, system cards, and release announcements},
  url={https://platform.claude.com/docs/en/about-claude/models/overview},
  note={Opus 4.6 system card February 2026; Sonnet 4.6 system card February 17, 2026. Accessed 2026-05-04.}
}

@misc{openai_gpt54_2026,
  title={{GPT-5.4} Thinking System Card},
  author={{OpenAI}},
  year={2026},
  howpublished={OpenAI system card, March 2026},
  url={https://openai.com/index/gpt-5-4-thinking-system-card/},
  note={Released March 5, 2026. Accessed 2026-05-04.}
}

@misc{openai_gpt54mini_2026,
  title={Introducing {GPT-5.4 mini} and nano},
  author={{OpenAI}},
  year={2026},
  howpublished={OpenAI release announcement},
  url={https://openai.com/index/introducing-gpt-5-4-mini-and-nano/},
  note={Accessed 2026-05-04.}
}

@misc{moonshot_kimi_k25_2026,
  title={{Kimi K2.5}: Open Visual Agentic Model for Real Work},
  author={{Moonshot AI}},
  year={2026},
  howpublished={Moonshot AI model documentation},
  url={https://www.kimi.com/ai-models/kimi-k2-5},
  note={Released January 27, 2026. Accessed 2026-05-04.}
}

@book{hutchins1995cognition,
  title={Cognition in the Wild},
  author={Hutchins, Edwin},
  year={1995},
  publisher={MIT Press},
  address={Cambridge, MA}
}

@article{clark1998extended,
  title={The Extended Mind},
  author={Clark, Andy and Chalmers, David},
  journal={Analysis},
  volume={58},
  number={1},
  pages={7--19},
  year={1998}
}

@article{wilson1927probable,
  title={Probable Inference, the Law of Succession, and Statistical Inference},
  author={Wilson, Edwin B.},
  journal={Journal of the American Statistical Association},
  volume={22},
  number={158},
  pages={209--212},
  year={1927},
  doi={10.1080/01621459.1927.10502953}
}

@misc{valsai_swev_2026,
  title={{SWE-Bench Verified}: Vals.ai Public Leaderboard (Bash-Tool-Only Harness)},
  author={{Vals AI}},
  year={2026},
  howpublished={Vals.ai third-party model evaluation leaderboard},
  url={https://www.vals.ai/benchmarks/swebench},
  note={Used for GPT-5.4 mini and GPT-5.4 xhigh SWE-Bench Verified scores because OpenAI does not publish vendor SWE-V numbers for the GPT-5.4 family. Vals.ai bash-tool-only harness scores: 73.0 for GPT-5.4 mini, 78.2 for GPT-5.4 xhigh. Accessed 2026-05-06.}
}

@misc{valsai_lcb_2026,
  title={{LiveCodeBench v6}: Vals.ai Public Leaderboard},
  author={{Vals AI}},
  year={2026},
  howpublished={Vals.ai third-party model evaluation leaderboard},
  url={https://www.vals.ai/benchmarks/lcb},
  note={Accessed 2026-05-06.}
}

@misc{valsai_terminalbench_2026,
  title={{Terminal-Bench 2.0}: Vals.ai Public Leaderboard},
  author={{Vals AI}},
  year={2026},
  howpublished={Vals.ai third-party model evaluation leaderboard},
  url={https://www.vals.ai/benchmarks/terminal-bench-2},
  note={Accessed 2026-05-06.}
}

\appendix
\section{Experimental details}
\label{app:experimental-details}

\subsection{Problem order and task structure}
Each esolang benchmark contains 80 tasks in a fixed fetch order
(20 easy, 20 medium, 20 hard, 20 extra-hard). Under the primary
protocol used for every main-text result, problems are fetched and
finalized in that fixed forward order: a problem is finalized either
when one hidden submission passes all six private tests or when the
three-submission cap is reached, after which the session advances to
the next problem and finalized problems are not revisited
(Figure~\ref{fig:methodology-pipeline} of the body). Workspace
isolation prevents the agent from reading sibling experiment folders
or any other run's artifacts.

\subsection{Task substrate details}
\label{app:task-details}
The four languages we evaluate are drawn from
EsoLang-Bench~\citep{esolangbench2026}, which releases task statements,
hidden tests, problem identifiers, difficulty tiers, and reference
interpreters under a single specification. The task statements,
hidden tests, and metadata used in this paper are loaded from the
Lossfunk EsoLang-Bench Hugging Face
dataset~\citep{esolangbench_hf_2026}; we reproduce only a brief
characterization of each language here so the appendix is
self-contained; the canonical problem list, per-tier difficulty
labels, and reference interpreters are documented in the
EsoLang-Bench release. The supplementary archive ships the four
interpreter Python sources used by the harness at
\texttt{supplementary\_code/benchmark\_harness/interpreters/} (one
file per language: Brainfuck, Befunge-98, Whitespace, Shakespeare),
and the public problem statements at
\texttt{benchmark\_harness/public/esolang\_full\_public.json}.

\textbf{Brainfuck.} Eight-symbol minimal pointer machine over an
unbounded tape of unsigned bytes. \texttt{>} \texttt{<} move the
pointer, \texttt{+} \texttt{-} mutate the cell with byte wrap,
\texttt{.} \texttt{,} do byte-level I/O, and \texttt{[} \texttt{]}
form a conditional loop on the current cell. There are no variables
or named functions; numeric I/O must be implemented digit by digit
through ASCII conversion. Programs in this language tend to be long,
fragile pointer-arithmetic sequences.

\textbf{Befunge-98.} Two-dimensional stack-based language whose
instruction pointer moves over a grid of one-character commands and
can be redirected by direction commands (\texttt{>} \texttt{<}
\texttt{\^{}} \texttt{v}), with stack arithmetic, string mode, an
end token (\texttt{@}), and grid get / put primitives. The 2D
control flow makes program structure positional rather than
sequential, and even small grid edits can change which path the
instruction pointer takes.

\textbf{Whitespace.} An assembly-style language whose lexicon
consists only of the three characters Space, Tab, and Linefeed.
Numbers are encoded as a sign bit followed by binary digits; control
flow uses labels and jumps. Because the source is invisible to a
human reader, programs must be produced by a generator script: any
direct edit through standard text-editing tools tends to silently
corrupt the program by altering whitespace.

\textbf{Shakespeare.} A natural-language shaped programming language
in which programs are written as theatrical scripts with named
characters and dialogue between exactly two on-stage characters at a
time. Statements set, read, print, and stack the listener's value,
expressions are built from positive and negative nouns scaled by
adjectives, and Roman-numeral scenes act as jump targets. The
syntax constraints are stylistic rather than minimal, but the
interpretive rules around speaker, listener, and stage state are
unfamiliar to typical software training data. EsoLang-Bench's fifth
language, Unlambda, is excluded due to interpreter latency
(Section~\ref{sec:eval}); the canonical problem list, hidden-test
contracts, and reference interpreters for the four languages we
keep are released by EsoLang-Bench.

\paragraph{Task set and tier structure.}
Each language ships $80$ problems split into four difficulty tiers of
$20$ each: easy (\texttt{E01}--\texttt{E20}), medium
(\texttt{M01}--\texttt{M20}), hard (\texttt{H01}--\texttt{H20}), and
extra-hard (\texttt{X01}--\texttt{X20}). The same problem statements
appear in every language, so cross-language differences come from the
target syntax and execution model rather than from problem
selection. The problems are short standard programming tasks that an
introductory programming course could pose; the difficulty in our
setting comes entirely from expressing them in an unfamiliar target
language. The easy tier covers I/O and one-step arithmetic (echo a
line, sum or multiply two integers, output a constant string,
character-by-character echo). The medium tier covers list and string
operations (sort a list of integers, count vowels, compute the
length of a string, parity / odd--even check, integer absolute
value, formatted multiplication tables, simple counters and
accumulators). The hard tier introduces multi-step numeric
manipulation (greatest common divisor, primality test, integer
division and modulo, Fibonacci, leading-zero suppression, signed
arithmetic with both inputs negative). The extra-hard tier exercises
combined control flow and data manipulation (least common multiple,
bracket-depth maximum, count inversions in a list, Roman-to-integer
conversion, base conversions, signed average / halving, string
rotation checks). EsoLang-Bench's own paper reports near-ceiling
zero-shot and few-shot accuracy on these same statements when models
answer in Python or JavaScript, which is the basis for our claim
that the difficulty in this benchmark comes from the target language
rather than the algorithmic problem. The full per-problem statements
ship in the supplementary archive at
\texttt{supplementary\_code/benchmark\_harness/public/esolang\_full\_public.json}.

\subsection{Harness commands}
\label{app:harness-commands}
The harness exposes a small command interface. \texttt{fetch} reveals the next
problem statement. \texttt{run} executes a candidate program locally with a
provided interpreter or verifier and user-specified input. \texttt{submit}
evaluates the candidate against private tests and updates the problem state.
\texttt{status} prints a progress dashboard and \texttt{export} dumps the full
per-cell session state as JSON. Agents can edit local files and run local
helper scripts inside their workspace. The full implementation is the single
file \texttt{supplementary\_code/benchmark\_harness/harness.py} in the
supplementary archive; the constant \texttt{MAX\_SUBMISSIONS\,=\,3} encodes the
hidden-submission cap and is enforced inside \texttt{harness.py} rather than by
the agent wrapper, so it cannot be bypassed.

\paragraph{Local execution limits.}
The four reference interpreters
(\texttt{supplementary\_code/\allowbreak benchmark\_harness/\allowbreak interpreters/})
terminate non-halting candidate programs by capping the number of executed
interpreter steps at \texttt{MAX\_INTERPRETER\_STEPS}\,=\,$10^7$ instructions
per local \texttt{run} (Brainfuck, Befunge-98, Whitespace, Shakespeare).
Programs that exceed the cap return a \texttt{StepLimitExceeded} runtime
error rather than hanging the session, which is important for Brainfuck
and Befunge-98 where infinite loops are easy to author. The same cap
applies inside hidden-test \texttt{submit} evaluation. The cap is
generous enough that none of the headline solutions reported in the
paper hit it; it functions as a watchdog, not as a difficulty knob.

\subsection{Primary protocol}
\label{app:primary-protocol}
The protocol used for every result in the main body allows unlimited
local interpreter calls and up to three hidden submissions per
problem. The agent decides when it has converged on the local
interpreter and then spends a hidden submission. The score is the
number of problems solved out of $80$. The full per-problem state
machine is in Figure~\ref{fig:methodology-pipeline} of the body, and
the operating parameters are summarized in
Table~\ref{tab:setup}.

\begin{table}[ht]
  \caption{\textbf{Primary protocol parameters.} All main-text
  results use this single configuration unless explicitly stated
  otherwise (the diagnostic variants in
  Section~\ref{sec:abl-meta}--\ref{sec:resources} relax or tighten
  individual rows but leave hidden tests, scoring, and workspace
  isolation unchanged).}
  \label{tab:setup}
  \centering
  \small
  \begin{tabular}{@{}lp{0.66\linewidth}@{}}
    \toprule
    Item & Value \\
    \midrule
    Languages & Brainfuck, Befunge-98, Whitespace, Shakespeare \\
    Problems per language & 80 ($20$ easy, $20$ medium, $20$ hard, $20$ extra-hard) \\
    Problem order & fixed forward; finalized problems not revisited \\
    Hidden tests per problem & 6 (private; agent never sees inputs or expected outputs) \\
    Hidden submissions per problem & up to 3 (\texttt{MAX\_SUBMISSIONS}\,=\,3 in \texttt{harness.py}) \\
    Local interpreter calls & unlimited (per-call step cap $10^7$ instructions) \\
    Output budget per assistant turn & $32{,}000$ tokens \\
    Sampling settings & provider / wrapper defaults (temperature, top-$p$, top-$k$) \\
    Workspace & isolated per model$\times$language run; sibling-folder access blocked \\
    Submission feedback & aggregate \texttt{passed/total} only; no per-test diagnostics \\
    Scoring rule & solved iff one submission returns $6/6$ on the hidden tests \\
    Aggregation across runs & headline cell = Session 1; two further sessions per cell as sanity checks (Appendix~\ref{app:per-session-results}); ablation cells = mean over two sessions (Appendix~\ref{app:robustness}) \\
    Uncertainty reporting & Wilson $95\%$ binomial CI~\citep{wilson1927probable} on $k/80$ \\
    \bottomrule
  \end{tabular}
\end{table}

\subsection{Per-agent API endpoints, model identifiers, and harness invocations}
\label{app:per-agent-apis}
Table~\ref{tab:agent-apis} lists the API endpoint, model identifier,
sampling configuration, and wrapper used for each of the six agents
in the headline runs. We do not override sampling at the API call
level: every agent runs at the wrapper's default temperature,
top-$p$, and top-$k$ settings, so the per-cell session-to-session
variation we report comes from independent re-invocations of the
wrapper rather than from explicit sampling changes. Example CLI
invocations: \texttt{claude --model claude-opus-4-6} (Claude Code),
\texttt{codex --model gpt-5.4 --reasoning xhigh} (Codex),
\texttt{opencode run --model moonshot/kimi-k2-thinking} (OpenCode);
the exact wrapper versions, environment variables, and per-cell
invocations used for the headline runs ship in the
\texttt{supplementary\_code/} archive.

\begin{table}[ht]
  \caption{Per-agent API endpoint, model identifier, sampling
  configuration, and harness invocation. ``Wrapper default''
  indicates that the agentic wrapper sets temperature, top-$p$, and
  top-$k$ to its built-in defaults; we do not override at the API
  call level. The model strings are the production identifiers used
  by the listed wrapper at the time of the headline runs.}
  \label{tab:agent-apis}
  \centering
  \footnotesize
  \setlength{\tabcolsep}{4pt}
  \begin{tabular}{@{}llllll@{}}
    \toprule
    Agent & Wrapper & Endpoint & Model identifier & Sampling & Reasoning effort \\
    \midrule
    Claude Opus 4.6   & Claude Code & Anthropic API & \texttt{claude-opus-4-6}    & wrapper default & --- \\
    Claude Sonnet 4.6 & Claude Code & Anthropic API & \texttt{claude-sonnet-4-6}  & wrapper default & --- \\
    Claude Haiku 4.5  & Claude Code & Anthropic API & \texttt{claude-haiku-4-5}   & wrapper default & --- \\
    GPT-5.4 xhigh     & Codex       & OpenAI API    & \texttt{gpt-5.4}            & wrapper default & xhigh   \\
    GPT-5.4 mini      & Codex       & OpenAI API    & \texttt{gpt-5.4-mini}       & wrapper default & medium  \\
    Kimi K2.5         & OpenCode    & Moonshot API  & \texttt{kimi-k2-thinking}   & wrapper default & ---     \\
    \bottomrule
  \end{tabular}
\end{table}

\subsection{Reporting and aggregation}
\label{app:reporting}
A problem is counted as solved only when all six private hidden tests
pass on a single submission. We do not aggregate per-test pass counts
across submissions, because partial-credit aggregation can inflate the
appearance of progress for runs whose submissions never fully pass. A
problem with three failed hidden submissions is counted as 0, not as
the best per-test count across the three. Cells that have not yet
completed all 80 problems are explicitly marked partial and excluded
from matched comparisons rather than imputed as zeros. Wilson 95\%
confidence intervals over solved counts are reported alongside the
headline cells in Section~\ref{app:uncertainty}.

\subsection{Robustness across independent sessions}
\label{app:robustness}

\paragraph{What a run means in this paper.}
We run \emph{independent sessions} of the underlying coding harness
rather than seeded re-inferences. Each headline
model$\times$language cell in Table~\ref{tab:main-results} is the
solved count from Session 1; we additionally ran two further
independent sessions per cell as session-to-session sanity checks
(per-session counts in Appendix~\ref{app:per-session-results}). Each
ablation cell (Section~\ref{sec:abl-meta},
Section~\ref{sec:distillation}, and Section~\ref{sec:resources}) is
the mean over two independent sessions. A session here is one fresh
end-to-end invocation of the agent's deployed CLI (Claude Code for
the Anthropic family, Codex for the GPT-5.4 family, OpenCode for
Kimi K2.5) on a freshly initialized workspace, with no state shared
across sessions.

\paragraph{Sessions, not seeds.}
The harnesses we evaluate are deployed CLI products and do not expose
a deterministic seed at the inference layer; the underlying provider
APIs do not return reproducible token sequences across requests at the
default temperatures and sampling configurations these CLIs use. Two
re-runs of the same model on the same problem set therefore differ at
the token level even with the same prompts, harness, and protocol
parameters. This captures more variation than a seeded re-run inside
a single inference call, because the spread also reflects session-level
choices: which helper file the agent writes first, whether it tries
direct target-language authoring before reaching for metaprogramming,
how aggressively it batches problems within the session, and which
debug-and-revise loops it falls into. We therefore use the terms
\emph{run} and \emph{session} consistently throughout the paper to
denote one such independent end-to-end CLI invocation, not a seeded
RNG call inside a single request.

\paragraph{Why three sessions per headline cell and two per ablation cell.}
For headline cells in Table~\ref{tab:main-results}, we report
Session 1 as the headline value and use Sessions 2 and 3 as
session-to-session sanity checks; per-session counts for all three
sessions are in Appendix~\ref{app:per-session-results}. Ablation
cells are paired contrasts on top of an established headline
(metaprogramming-allowed vs.~direct authoring in
Section~\ref{sec:abl-meta}, with vs.~without a reference library in
Section~\ref{sec:distillation}, and varying interpreter-call and
output-token budgets in Section~\ref{sec:resources}); the relevant
signal is the within-agent, within-language shift induced by the
intervention rather than the absolute score, so we ran two
independent sessions per ablation cell and report the mean. In
every reported ablation, the qualitative direction of the
intervention is consistent across both sessions.

\paragraph{What this means for the headline claims.}
The inter-agent separations that drive Section~\ref{sec:cliff} and
Section~\ref{sec:abl-meta} of the body are tens of problems out of
$80$ (for example $79$ vs.~$4$ on Brainfuck and $80$ vs.~$4$ on
Befunge-98). Observed session-to-session variation across the three
sessions per headline cell is small relative to these gaps
(Appendix~\ref{app:per-session-results}), so the agent ordering
reported in the body is stable under independent re-runs. The
per-cell binomial Wilson interval on the solved-out-of-$80$ counts
is in Appendix~\ref{app:uncertainty}.

\subsection{Controlled-access (fixed-budget) protocol}
\label{app:fixed-budget}
For a stricter reproducibility-oriented control we also ran a
controlled-access protocol that caps local runs and hidden submissions
explicitly. Each problem allows at most three local interpreter or
verifier calls before a single hidden submission is unlocked (the
agent may submit sooner if it is confident, but cannot make a fourth
local run before submitting), and at most one hidden submission per
problem. Hidden tests, scoring rule, problem order, and workspace
isolation are unchanged from the primary protocol; only the local-run
and submission caps are tightened. The qualitative agent ordering
under this stricter protocol matches the primary protocol; absolute
scores are lower across the board because frontier agents lose access
to the iterative repair loop they normally use. Per-cell numbers are
recorded in the supplementary CSV.

\subsection{Interpreter-budget ablation}
\label{app:interpreter-budget}
The interpreter-budget ablation
(Section~\ref{sec:abl-budget}, Figure~\ref{fig:scaling-curve} of the
body) caps the number of local interpreter calls per problem at
$\{3, 5, 15, 30, \infty\}$ while holding the task substrate, hidden
submissions, and scoring rule fixed. The full per-budget solved
counts on both Brainfuck and Befunge-98 are the values plotted in
body Figure~\ref{fig:scaling-curve}; the unlimited-budget endpoints
match the headline cells in Table~\ref{tab:main-results} of the body
($64$ for Opus 4.6 on each of Brainfuck and Befunge-98, $12$ for
Sonnet 4.6 on Brainfuck and $64$ on Befunge-98, $4$ for Haiku 4.5 on
each). The qualitative pattern reported in the body is that Opus 4.6
gains substantially as the budget grows on both languages and Sonnet
4.6 gains on Befunge-98, while Haiku 4.5 stays near the floor at
every budget. Per-cell raw counts are emitted by the per-cell
\texttt{export.json} files under
\texttt{supplementary\_code/experiments/01\_main\_experiments/} and
its budget-restricted variants.

\subsection{Metaprogramming constraints}
\label{app:meta-constraints}
In metaprogramming-allowed runs, the agent may write generator programs in a
familiar language that emit the target esolang. In no-meta runs, the agent must
write the target language directly. Cross-language generator runs constrain the
generator language to Python, JavaScript, or Rust depending on the cell.

\subsection{Isolation}
\label{app:isolation}
Each run receives its own workspace. Agents are instructed not to inspect prior
runs, sibling language folders, hidden tests, solved artifacts, or transcripts.
This isolation is important because the benchmark is sequential: later problems
can legitimately reuse notes and primitives developed earlier in the same run,
but not artifacts from other runs.

\subsection{System prompts and per-condition configuration}
\label{app:system-prompts}
Each cell of the paper runs under one of three agentic harnesses
(Claude Code, Codex, OpenCode), each of which automatically reads a
project-level instruction file from the workspace at session start
(\texttt{CLAUDE.md} for Claude Code, \texttt{AGENTS.md} for Codex and
OpenCode) on top of its own internal default system prompt. We ship
one such instruction file per language; its content is what the
harness adds to its native default. The four prompts reproduced
verbatim in the subsections below are the exact files shipped in the
supplementary archive at \texttt{supplementary\_code/prompts/<lang>/}
(symlinked into every primary-protocol cell at
\texttt{experiments/01\_main\_experiments/<harness>/<model>/<lang>/}).
The harness's own internal
default (turn-taking, tool-use formatting, file-editing rules) is
left unchanged. We refer to the file we ship as the
\emph{language-reference prompt}.

The language-reference prompt has the same structure for every
esolang: a one-line ``start solving now'' directive, the
benchmark task description, the harness command list
(\texttt{fetch}, \texttt{run}, \texttt{submit}, \texttt{status},
\texttt{skip}), the scoring rule (a problem is solved only if all
six private hidden tests pass), the operating limits (up to three
hidden submissions per problem, unlimited \texttt{run} calls,
unlimited time, no skipping with submissions remaining), and a
language-specific reference card. This single
language-reference prompt is identical across the three Claude
agents (Opus 4.6, Sonnet 4.6, Haiku 4.5) within a given language;
for Codex and OpenCode we use the same content via the
\texttt{AGENTS.md} mechanism. The four primary-protocol prompts
are reproduced verbatim in
Sections~\ref{app:prompt-brainfuck}--\ref{app:prompt-shakespeare}
below.

Per-condition deviations from this primary-protocol prompt are as
follows.

\textbf{Primary protocol and main results
(Section~\ref{sec:results}).} The language-reference prompt
described above (verbatim text in
Sections~\ref{app:prompt-brainfuck}--\ref{app:prompt-shakespeare}).
No further additions. The agent receives each problem statement
through \texttt{fetch}; everything else is handled by the harness.

\textbf{Metaprogramming-allowed (Section~\ref{sec:abl-meta}).}
Same as the primary protocol above (the language-reference prompt is loaded as
\texttt{CLAUDE.md} or \texttt{AGENTS.md}, on top of the harness's native default).
The agent freely chooses whether to author the target esolang directly
or via a host-language generator; no additional preamble is added.

\textbf{No-meta direct authoring (Section~\ref{sec:abl-meta}).}
The harness removes the agent's bash access and any tool that
could execute a host-language generator, and a short
paper-authored preamble is added to the system prompt instructing
the agent to author the target esolang directly. The preamble
lists the file extensions accepted by the harness for that cell
(e.g., \texttt{.bf} only) and explicitly notes that any generator
script in the workspace at submission time will cause submission
failure. Together, the bash removal and the prompt preamble
prevent any host-language generation pathway.

\textbf{Cross-language generator transfer
(Section~\ref{sec:abl-meta}).} A short preamble specifies which
host language is permitted (Python, JavaScript, or Rust per cell)
and lists the file extensions accepted for the generator. All other
harness behavior is unchanged.

\textbf{Distillation: text-only preamble (Section~\ref{sec:distillation}, condition (i)).}
A condensed prose version of Opus 4.6's strategies is preloaded as
the system prompt. The full text is reproduced in
Appendix~\ref{app:distillation-text-prompt}.

\textbf{Distillation: reference library (Section~\ref{sec:distillation}, condition (ii)).}
The condition (i) preamble plus a \texttt{reference\_lib/} directory
containing a small strategy-only library distilled from the
strong-agent traces (Claude Opus 4.6, with one generic Brainfuck
builder pattern from GPT-5.4 xhigh's session). The preamble points
the agent at the reference directory and its \texttt{README.md}.
The library contains generic code-generation primitives only; no
per-problem generators and no solved benchmark programs are
included. The exact files shipped are listed in
Appendix~\ref{app:distillation-reflib-prompt}, and the full preamble
text is reproduced there.

\textbf{Interpreter-budget ablation
(Section~\ref{sec:abl-budget}).} Harness defaults plus a hard cap
on local interpreter calls per problem (3, 5, 15, 30, or
unlimited). The cap is enforced by the harness rather than the
prompt.

\textbf{Token-efficiency ablation (Section~\ref{sec:abl-tokens}).}
Harness defaults; the cell is read out of the primary-protocol
runs by counting API output tokens (including extended thinking)
on E01--E20 only.

\subsubsection{Primary-protocol language-reference prompt: Brainfuck}
\label{app:prompt-brainfuck}

\begin{footnotesize}
\begin{verbatim}
> **Start solving now.** The session is initialized. Begin by fetching your first problem:
> ```
> python harness.py fetch
> ```
> Work through all 80 problems sequentially. For each problem: read
> the description, generate your own test cases, use `run` to test,
> then `submit` when confident. Do NOT skip without attempting --
> see skip policy below.

# EsoLang-Bench: Brainfuck

You are solving 80 brainfuck programming problems sequentially.
Problems span 4 difficulty tiers: E01-E20 (easy), M01-M20 (medium),
H01-H20 (hard), X01-X20 (extra-hard).

Each problem has 6 hidden test cases you cannot see. No test cases
are provided -- generate your own test inputs from the problem
descriptions.

## Harness Commands

```bash
python harness.py fetch                          # Get next problem
python harness.py run <code_file> --input "..."  # Test with custom input (unlimited)
python harness.py submit <problem_id> <file>     # Hidden evaluation (max 3 per problem)
python harness.py status                         # Progress dashboard
python harness.py skip                           # Skip ONLY after attempting (see policy)
```

## Rules

- Solve problems in order. `fetch` advances to the next.
- Maximum 3 submissions per problem. Use `run` (unlimited) to test first.
- `submit` returns PASS/WRONG ANSWER/RUNTIME ERROR per hidden test
  -- no inputs or expected outputs shown.
- Problem is solved only if 6/6 hidden tests pass.
- Do not read private test files, reverse-engineer the harness, or
  hardcode outputs.

## CRITICAL: No Skipping -- Use All 3 Submissions

- You MUST work on every problem until you either score 6/6 OR use
  all 3 submissions. Do NOT move to the next problem with unused
  submissions remaining.
- You CANNOT revisit problems later -- once you move on, the problem
  is gone forever. Every unused submission is wasted.
- There is NO time limit. You have UNLIMITED time and UNLIMITED test
  runs (`run` command). The ONLY limit is 3 submissions per problem.
- For each problem: read the description, write a genuine solution,
  test it thoroughly with `run` using multiple inputs until it
  produces correct output, THEN submit.
- Do NOT submit untested code. Do NOT submit placeholder or stub
  solutions. Every submission must be a genuine, tested attempt.
- If a submission returns WRONG ANSWER or RUNTIME ERROR: analyze the
  failure, try a completely different approach, test it, then use
  your next submission.
- Even if a problem seems extremely difficult, you must use all 3
  submissions with genuinely different approaches before moving on.
  A partial score (1-5/6) is far better than 0/6.
- Do NOT call `fetch` or `skip` to move to the next problem while
  you still have submissions remaining on the current one.

## Scoring

- Per problem: tests passed / 6
- Overall: problems solved / 80 and total tests passed / 480

## Brainfuck Language Reference

### Commands

| Cmd | Description |
|-----|-------------|
| `>` | Move pointer right |
| `<` | Move pointer left |
| `+` | Increment cell (wraps 255->0) |
| `-` | Decrement cell (wraps 0->255) |
| `.` | Output cell as ASCII character |
| `,` | Read one byte into cell (0 if EOF) |
| `[` | Jump past matching `]` if cell is 0 |
| `]` | Jump back to matching `[` if cell nonzero |

All other characters are comments.

### Memory Model

- Unbounded tape of unsigned byte cells (0-255), initialized to 0.
- Pointer starts at cell 0; cannot go below 0 (runtime error).

### Essential Patterns

- Zero a cell:        [-]
- Move (destructive): [->+<] (cell 0 to cell 1)
- Copy:               [->+>+<<]>>[-<<+>>]
- Add cell 1 into 0:  >[-<+>]<
- Subtract 1 from 0:  >[-<->]<
- Read loop:          ,[...,] reads until EOF (0)
- Print decimal:      Divide by 10, store remainders, print in
                      reverse + ASCII 48

### ASCII Quick Reference

`0`=48, `9`=57, `A`=65, `Z`=90, `a`=97, `z`=122, space=32,
newline=10, `-`=45

### Common Pitfalls

- Number I/O: Input arrives as ASCII chars ('5'=53), not raw
  numbers. Output must also be ASCII digits. This is the #1 error
  source.
- Forgetting to zero cells before reuse.
- Cell overflow: 255+1=0, 0-1=255. Can cause infinite loops.
- Pointer tracking: Always keep a written map of which cell holds
  what.
- Multi-digit numbers: Parsing and printing require digit-by-digit
  handling.
\end{verbatim}
\end{footnotesize}

\subsubsection{Primary-protocol language-reference prompt: Befunge-98}
\label{app:prompt-befunge}

The header, harness commands, rules, no-skipping policy, and
scoring section are identical to the Brainfuck prompt above (with
``brainfuck'' replaced by ``befunge-98''). Only the language
reference card differs and is reproduced below.

\begin{small}
\begin{verbatim}
## Befunge-98 Language Reference

### Program Structure

2D grid. Instruction pointer starts at (0,0) moving right. Wraps at
edges. Ends at `@`.

### Instructions

Stack:    0-9 push digit; a-f push 10-15; " toggle string mode;
          : dup; \ swap; $ pop; n clear stack
Arith:    + - * / %  (pop b, pop a, push a op b)
Compare:  ! (NOT); ` (greater-than)
Direction: > < ^ v ?(random) [(turn left) ](turn right) r(reverse)
           #(skip next) j(jump n)
Branch:   _ (right if 0, left if nonzero); | (down if 0, up if
          nonzero)
I/O:      . print int+space; , print char; & read int;
          ~ read char (-1 at EOF)
Grid:     g get; p put; s store next; ' fetch next
Flow:     @ end; q quit; ; comment toggle; space=nop

### Essential Patterns

- Print string: "!dlroW olleH">:#,_@ (push reversed, loop print)
- Print loop:   >:#,_@ (dup, skip print if 0, print char, repeat)
- Read int:     & (built-in)
- Conditional:  !#v_ (branch on value)

### Common Pitfalls

- `.` outputs number + space. For clean output, convert to digit
  chars and use `,`.
- Stack order: 52- = 5-2=3, not 2-5.
- String mode: "abc" pushes a,b,c in order; they pop as c,b,a.
- Missing `@`: IP wraps and re-executes, hitting step limit.
- `~` returns -1 at EOF, not 0.
\end{verbatim}
\end{small}

\subsubsection{Primary-protocol language-reference prompt: Whitespace}
\label{app:prompt-whitespace}

The header, harness commands, rules, no-skipping policy, and
scoring section are identical to the Brainfuck prompt above (with
``brainfuck'' replaced by ``whitespace''). Only the language
reference card differs and is reproduced below.

\begin{small}
\begin{verbatim}
## Whitespace Language Reference

### Encoding

Programs use only: Space (S, ASCII 32), Tab (T, ASCII 9), Linefeed
(L, ASCII 10). All other chars are ignored.

### Instruction Encoding (selected)

Stack:    SS<num>  push number; SLS dup; SLT swap; SLL discard
Arith:    TSSS add; TSST sub; TSSL mul; TSTS div; TSTT mod
Heap:     TTS store; TTT retrieve
I/O:      TLSS out_char; TLST out_num; TLTS read_char; TLTT read_num
Flow:     LSS<lbl> label; LST<lbl> call; LSL<lbl> jump;
          LTS<lbl> jz; LTT<lbl> jn (jump if negative);
          LTL ret; LLL end

### Key Concepts

- Number encoding: Sign bit (S=+, T=-) + binary digits
  (S=0, T=1) + L terminator.
- Heap-based I/O: read_char/read_num store to heap at a popped
  address, not the stack. Push address first, then read, then
  push address + retrieve to get value onto stack.
- Labels: Sequences of S and T terminated by L. Must be unique.

### Common Pitfalls

- Heap I/O: Read operations store to heap, not stack. Must
  retrieve afterward.
- Sign bit required even for positive numbers.
- Always include the end instruction (LLL) or execution runs
  off the end.
- Stack order: sub pops b then a, computes a-b.
\end{verbatim}
\end{small}

\subsubsection{Primary-protocol language-reference prompt: Shakespeare}
\label{app:prompt-shakespeare}

The header, harness commands, rules, no-skipping policy, and
scoring section are identical to the Brainfuck prompt above (with
``brainfuck'' replaced by ``shakespeare''). Only the language
reference card differs and is reproduced below.

\begin{small}
\begin{verbatim}
## Shakespeare Language Reference

### Program Structure

Title.
Character, description.
Act I: Description.
Scene I: Description.
[Enter Character1 and Character2]
Character1:
Statement.

### Value System

- Positive nouns (+1): angel, cat, day, flower, hero, joy, king,
  rose, summer, sun
- Negative nouns (-1): bastard, beast, coward, death, devil,
  famine, hell, pig, plague
- Zero: nothing, zero
- Each adjective doubles: "a big cat"=2, "a big big cat"=4,
  "a big big big cat"=8
- Arithmetic: the sum of X and Y, the difference between X and Y,
  the product of X and Y, the quotient between X and Y, the
  remainder of the quotient between X and Y, the square of X,
  twice X
- Pronouns: you / thou / thee = listener's value;
            I / me           = speaker's value

### Statements (all target the LISTENER)

- Assign:        You are EXPR. / Thou art EXPR.
- Output char:   Speak your mind.        (listener's value as char)
- Output number: Open your heart.        (listener's value as int)
- Read char:     Open your mind.         (-1 at EOF)
- Read int:      Listen to your heart.
- Stack push:    Remember EXPR.          (onto listener's stack)
- Stack pop:     Recall.                 (pop listener's stack into
                                          listener's value)

### Comparisons and Flow Control

- Am I better than EXPR?  (speaker > EXPR)
- Am I worse than EXPR?   (speaker < EXPR)
- Am I as good as EXPR?   (speaker == EXPR)
- If so, let us proceed to Scene X.   (jump if true)
- If not, let us proceed to Scene X.  (jump if false)
- Let us proceed to Scene X.          (unconditional jump)

### Stage Rules

- Exactly 2 characters on stage for dialogue.
- [Enter X and Y], [Exit X], [Exeunt] (remove all)
- Scene labels use Roman numerals and must be globally unique.

### Common Pitfalls

- I/O targets the LISTENER, not the speaker. This is the #1 bug
  source.
- No direct numbers: "You are 72." is INVALID. Use adjective-noun
  expressions.
- Stage management: Must have exactly 2 characters on stage.
- Roman numeral scenes must be globally unique across all Acts.
\end{verbatim}
\end{small}

\section{Additional results}
\label{app:additional-results}

Every per-cell number reported in this appendix is reproducible from
the per-cell \texttt{export.json} files emitted by
\texttt{python harness.py export}; the supplementary archive ships
the harness, the 48 ready-to-run cell directories, and a rigorous
end-to-end test (\texttt{scripts/rigorous\_test.sh}) that exercises
the export path without requiring any provider API key.

\subsection{Terminal-Bench 2.0 and SWE-Bench Verified vs
EsoLang-Bench scatter (cliff visualization)}
\label{app:cliff-scatter}
\begin{figure}[ht]
  \centering
  \includegraphics[width=\linewidth]{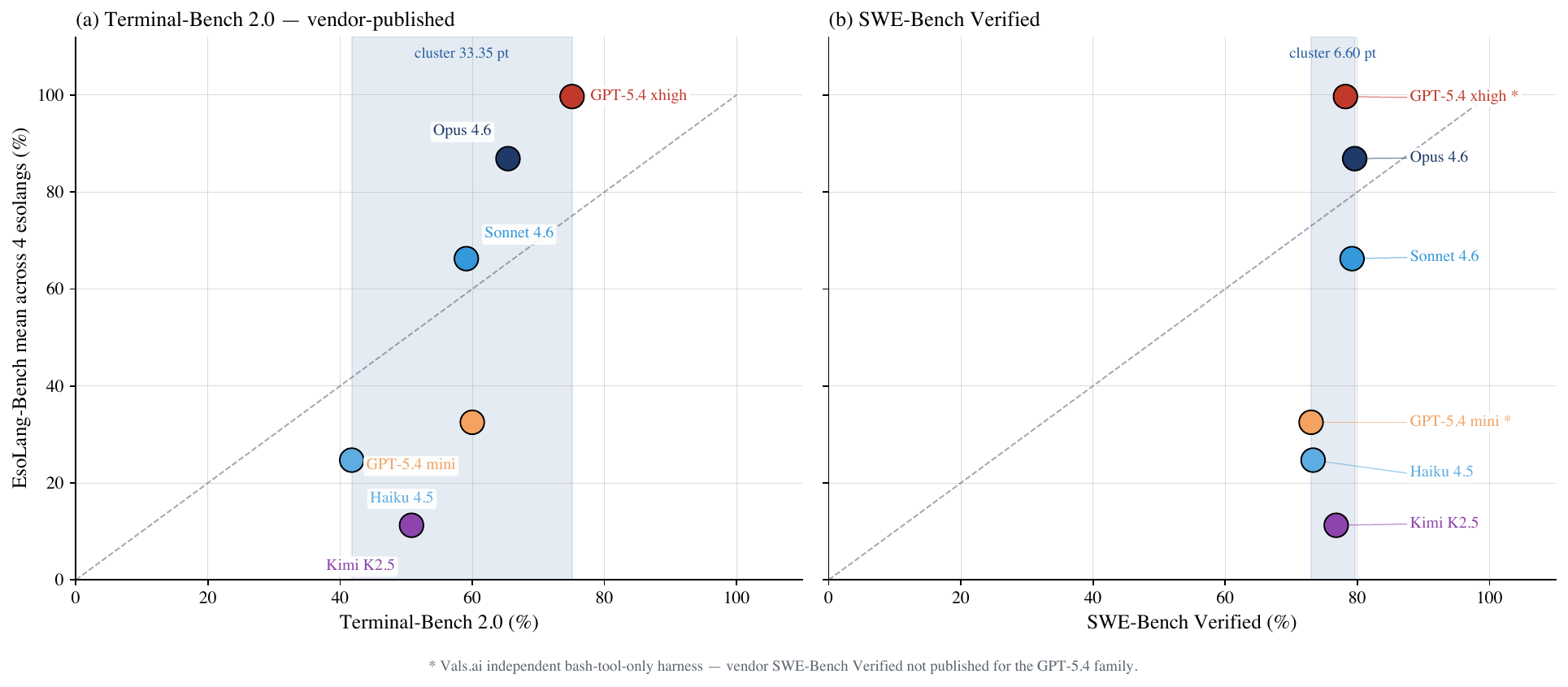}
  \caption{\textbf{Mainstream coding scores cluster while
  unfamiliar-language scores separate, on both Terminal-Bench 2.0
  and SWE-Bench Verified.} Each marker is one of the six evaluated
  coding agents. (a) Terminal-Bench 2.0 (vendor-published) on the
  x-axis; mean EsoLang-Bench score under our protocol on the y-axis;
  shaded vertical band is the $33.3$-pt TB-2.0 cluster. (b) SWE-Bench
  Verified on the x-axis with the same y-axis; shaded vertical band is
  the $6.6$-pt SWE-V cluster. Asterisks mark Vals.ai
  bash-tool-only-harness numbers used where vendor SWE-V scores are
  not published for the GPT-5.4 family.}
  \label{fig:cliff-scatter}
\end{figure}

Figure~\ref{fig:cliff-scatter} below visualizes two columns of
Table~\ref{tab:cliff-spread} of the body, Terminal-Bench 2.0 and
SWE-Bench Verified, against each agent's EsoLang-Bench mean score.
The shaded vertical band in each panel is the agent cluster on the
mainstream benchmark ($33.3$ pp wide on Terminal-Bench 2.0 and
$6.6$ pp wide on SWE-Bench Verified), while the vertical extent of
the markers shows the $88.4$-pt EsoLang-Bench spread of the same
six agents.

\subsection{Main results in raw problems-solved counts}
\label{app:main-results-raw}
Table~\ref{tab:main-results} in the body reports the four esolang
columns as percentages out of 80 problems per language for
readability. Table~\ref{tab:main-results-raw} reproduces the same
six-agent results in the underlying problems-solved-out-of-80 format.

\paragraph{Mainstream-benchmark sourcing.}
\label{app:swev-sourcing}
The SWE-Bench Verified and Terminal-Bench 2.0 rows of
Table~\ref{tab:cliff-spread} use vendor-published numbers wherever
those exist. We did not re-run SWE-Bench Verified or Terminal-Bench
2.0 ourselves for the headline numbers; every cell in those two rows
is sourced from the public reports listed in
Table~\ref{tab:swev-sourcing} below. Where no vendor SWE-V score is
published, we use the Vals.ai third-party
leaderboard~\citep{valsai_swev_2026}, which evaluates SWE-Bench
Verified under a published bash-tool-only harness; this applies only
to the GPT-5.4 family. As a sanity check on the third-party numbers,
we additionally re-ran SWE-Bench Verified on GPT-5.4 mini and
GPT-5.4 xhigh under our own harness and recovered scores within a
few points of the Vals.ai cells, so the headline values are stable
under independent replication; we use the Vals.ai numbers in the
table for transparency about the source.

\begin{table}[ht]
  \caption{\textbf{Per-agent sourcing for the SWE-Bench Verified and
  Terminal-Bench 2.0 rows of Table~\ref{tab:cliff-spread}.} Every
  cell is a public report from the listed source; we did not re-score
  any of the mainstream cells ourselves for the headline tables. The
  Vals.ai entries are used only because OpenAI does not publish
  vendor SWE-V numbers for the GPT-5.4 family.}
  \label{tab:swev-sourcing}
  \centering
  \footnotesize
  \begin{tabular}{@{}llll@{}}
    \toprule
    Agent & SWE-V (\%) & TB-2.0 (\%) & Source \\
    \midrule
    Claude Opus 4.6   & $79.6$ & $65.4$ & Anthropic system card / model overview~\citep{anthropic_models_2026} \\
    Claude Sonnet 4.6 & $79.2$ & $59.1$ & Anthropic system card / model overview~\citep{anthropic_models_2026} \\
    Claude Haiku 4.5  & $73.3$ & $41.8$ & Anthropic system card / model overview~\citep{anthropic_models_2026} \\
    GPT-5.4 xhigh     & $78.2$ & $75.1$ & SWE-V: Vals.ai~\citep{valsai_swev_2026}; TB-2.0: OpenAI~\citep{openai_gpt54_2026} \\
    GPT-5.4 mini      & $73.0$ & $60.0$ & SWE-V: Vals.ai~\citep{valsai_swev_2026}; TB-2.0: OpenAI~\citep{openai_gpt54mini_2026} \\
    Kimi K2.5         & $76.8$ & $50.8$ & Moonshot release notes / model card~\citep{moonshot_kimi_k25_2026} \\
    \bottomrule
  \end{tabular}
\end{table}

\begin{table}[ht]
  \caption{Main esolang results in raw problems-solved-out-of-80
  format. Same data as the percentage columns of
  Table~\ref{tab:main-results}; multiply each cell by $100/80$ to
  recover the percentage.}
  \label{tab:main-results-raw}
  \centering
  \small
  \begin{tabular}{lcccc}
    \toprule
    Model & Whitespace & Shakespeare & Befunge-98 & Brainfuck \\
    \midrule
    Kimi K2.5 & 25 & 2 & 5 & 4 \\
    Claude Haiku 4.5 & 65 & 6 & 4 & 4 \\
    Claude Sonnet 4.6 & 80 & 56 & 64 & 12 \\
    Claude Opus 4.6 & 80 & 70 & 64 & 64 \\
    GPT-5.4 mini & 71 & 17 & 11 & 5 \\
    GPT-5.4 xhigh & 80 & 80 & 80 & 79 \\
    \bottomrule
  \end{tabular}
\end{table}

\subsection{Opus 4.6 Brainfuck local-call distribution at budget $30$}
\label{app:budget-30}
This subsection breaks down the per-problem distribution of local
interpreter calls used \emph{within the budget-$30$ cell of
Section~\ref{sec:abl-budget}}, conditional on the problem being
solved in that cell. It is descriptive only: the headline budget-$3$
cell of Figure~\ref{fig:scaling-curve} is a separate run with a
different cap, and a problem solved here at, say, call $8$ would not
have been reached in a budget-$3$ run because the agent could not
have afforded the eighth call there. The distribution should
therefore not be read as predicting the budget-$3$ score.

Among the $k$ solved problems in the budget-$30$ Opus 4.6 Brainfuck
cell, the local-call distribution observed at solve time is: a
plurality solve within the first three local calls, with the
remainder spread across higher call counts (one or two solves each
at calls $4$--$9$, and isolated solves at calls $12$, $13$, and
$19$). The exact per-problem counts are recorded in the
budget-$30$ \texttt{export.json} for that cell; the body figure
plots the per-cell aggregates, not this within-cell distribution.

\subsection{Cross-harness evidence}
\label{app:cross-harness}
The headline comparison is already cross-harness by construction.
The three model families we evaluate run under three independently
implemented agentic wrappers: Claude Opus 4.6, Sonnet 4.6, and Haiku
4.5 under \emph{Claude Code} (Anthropic); GPT-5.4 xhigh and GPT-5.4
mini under \emph{Codex} (OpenAI); Kimi K2.5 under \emph{OpenCode}
(Moonshot, third-party). These three wrappers ship different default
system prompts, different file-editing semantics, different
turn-taking conventions, and different shell-tool surfaces. We hold
the benchmark-facing operations constant across them
(\texttt{fetch}, \texttt{run}, \texttt{submit}, \texttt{status}) and
ship the same per-language language-reference prompt (the
\texttt{CLAUDE.md} text reproduced in
Sections~\ref{app:prompt-brainfuck}--\ref{app:prompt-shakespeare},
loaded as \texttt{CLAUDE.md} under Claude Code and as
\texttt{AGENTS.md} under Codex and OpenCode). The same capability
ordering and the same large per-language separation between frontier
and weaker agents (Table~\ref{tab:main-results}) appear on top of
all three independent wrappers, so the order-of-magnitude headline
spread is not a single-wrapper artifact.

Two additional consistency checks support this interpretation. First,
within the Claude family running entirely under Claude Code, the
$80$-point gap between Opus 4.6 and Haiku 4.5 on Brainfuck and
Befunge-98 is too large to be explained by per-wrapper tooling
differences, since wrapper and harness are held constant within the
family. Second, on the languages where direct authoring is feasible
(Whitespace and Shakespeare for the strongest two agents) the
between-family ordering and absolute scores under three different
wrappers fall within a narrow band, indicating that wrapper
differences are second-order relative to the capability differences
the headline numbers reveal.

\paragraph{Single-model-multiple-wrappers OpenCode check.}
On top of this natural three-wrapper diversity we ran an explicit
single-model-multiple-wrappers control: we re-ran the strongest
agent in each of the Claude and GPT-5.4 families
(Opus~4.6 and GPT-5.4~xhigh) under \emph{OpenCode} on the two
diagnostic languages where direct authoring is most fragile and
where the body's metaprogramming finding is most diagnostic
(Brainfuck and Befunge-98). All other protocol parameters are held
fixed: the same EsoLang-Bench task statements, the same six private
hidden tests per problem, the same up-to-three-submissions cap, the
same unlimited local interpreter access, and the same per-language
language-reference prompt loaded as \texttt{AGENTS.md}.

\begin{table}[ht]
  \caption{\textbf{Single-model-multiple-wrappers OpenCode check.}
  Solved problems out of 80 for the strongest agent in each native
  family on the two diagnostic languages, comparing the native
  wrapper to OpenCode while holding everything else fixed. Each
  one-problem step on this scale is $1.25$ percentage points; both
  models drop by 1--2 problems out of 80 under OpenCode, well within
  the Wilson $95\%$ binomial CI of the native count, and far below
  the 50--80-problem separations between frontier and weaker agents
  in the headline cells.}
  \label{tab:cross-harness-opencode}
  \centering
  \small
  \begin{tabular}{llccc}
    \toprule
    Model & Native wrapper & Language & Native (k/80) & OpenCode (k/80) \\
    \midrule
    Claude Opus 4.6  & Claude Code & Brainfuck   & 64 & 62 \\
    Claude Opus 4.6  & Claude Code & Befunge-98  & 64 & 63 \\
    GPT-5.4 xhigh    & Codex       & Brainfuck   & 79 & 77 \\
    GPT-5.4 xhigh    & Codex       & Befunge-98  & 80 & 78 \\
    \bottomrule
  \end{tabular}
\end{table}

The native-versus-OpenCode delta is at most $-2$ problems out of 80
in every cell. For the two strongest agents on the two diagnostic
languages, the OpenCode re-run lands within the Wilson $95\%$ CI of
the native-wrapper count and preserves the qualitative ordering
(Opus~4.6 strong on Brainfuck, near-ceiling on Befunge-98;
GPT-5.4~xhigh near-ceiling on both). Combined with the natural
three-wrapper diversity of the headline comparison, this confirms
that the headline ordering and the order-of-magnitude per-language
separation between frontier and weaker agents are not artifacts of
Claude Code or Codex specifically.

\subsection{Kimi K2.5 under OpenCode}
\label{app:kimi-runs}
Kimi K2.5 is run under OpenCode in the headline cells because the
Anthropic Claude Code wrapper does not load Kimi K2.5 as a native
target and the OpenAI Codex wrapper does not host third-party
checkpoints under its function-calling protocol. The OpenCode runs
use the same harness command interface (\texttt{fetch}, \texttt{run},
\texttt{submit}) and the same language-reference prompt
(\texttt{AGENTS.md}) as Codex. The four headline Kimi K2.5 cells
solve $4/80$ on Brainfuck, $5/80$ on Befunge-98, $25/80$ on
Whitespace, and $2/80$ on Shakespeare; these are reproduced as
percentages in Table~\ref{tab:main-results} of the main text and as
raw counts in Table~\ref{tab:main-results-raw}. The headline cells
already use the primary metaprogramming-allowed protocol (the
agent is free to write a host-language generator if it chooses), so
no separate ``meta-allowed'' Kimi K2.5 variant is reported: the
headline numbers \emph{are} the meta-allowed numbers, and they place
Kimi K2.5 in the low-performance regime that the body identifies as
capability-gated.

\subsection{Confidence intervals}
\label{app:uncertainty}
\label{app:wilson-asym}
Each headline cell in the main results is the \textbf{Session 1}
solved count (Appendix~\ref{app:robustness}). The reported
uncertainty for this Session 1 count is a $95\%$ \textbf{Wilson
binomial confidence interval (Wilson CI)}~\citep{wilson1927probable}
on the raw count $k/80$ of that session; a bootstrap over the $80$
per-problem outcomes ($10{,}000$ resamples) produces quantitatively
similar intervals. Table~\ref{tab:main-results} in the body reports
the worst-side Wilson CI half-width as a single $\pm$ subscript for
readability; the full asymmetric Wilson CIs are reproduced in
Table~\ref{tab:main-results-asym} below. Each cell shows percentage
solved with a separate upper and lower half-width, so that bounded
cells (near $0\%$ or $100\%$) are visible as one-sided or
near-one-sided. Half-widths are computed from the Wilson score
formula on raw counts $k/80$ for per-language cells and $k/320$ for
the pooled \textbf{EsoLang mean}.

\begin{table}[ht]
  \caption{Asymmetric Wilson 95\% binomial confidence intervals on
  the headline cells of Table~\ref{tab:main-results}.}
  \label{tab:main-results-asym}
  \centering
  \footnotesize
  \setlength{\tabcolsep}{3.5pt}
  \begin{tabular}{lccccc}
    \toprule
    Model & WS & Sh & B98 & BF & \textbf{EsoLang mean} \\
    \midrule
    Kimi K2.5         & $31.3^{+10.8}_{-9.1}$ & $2.5^{+6.2}_{-1.8}$    & $6.3^{+7.6}_{-3.6}$    & $5.0^{+7.2}_{-3.0}$    & $\mathbf{11.3^{+3.9}_{-3.0}}$ \\
    Claude Haiku 4.5  & $81.3^{+7.0}_{-9.9}$  & $7.5^{+7.9}_{-4.0}$    & $5.0^{+7.2}_{-3.0}$    & $5.0^{+7.2}_{-3.0}$    & $\mathbf{24.7^{+5.0}_{-4.4}}$ \\
    Claude Sonnet 4.6 & $100.0^{+0.0}_{-4.6}$ & $70.0^{+8.9}_{-10.8}$  & $80.0^{+7.3}_{-10.0}$  & $15.0^{+9.4}_{-6.2}$   & $\mathbf{66.3^{+5.0}_{-5.3}}$ \\
    Claude Opus 4.6   & $100.0^{+0.0}_{-4.6}$ & $87.5^{+5.6}_{-9.0}$   & $80.0^{+7.3}_{-10.0}$  & $80.0^{+7.3}_{-10.0}$  & $\mathbf{86.9^{+3.3}_{-4.1}}$ \\
    GPT-5.4 mini      & $88.8^{+5.2}_{-8.8}$  & $21.3^{+10.2}_{-7.5}$  & $13.8^{+9.2}_{-5.9}$   & $6.3^{+7.6}_{-3.6}$    & $\mathbf{32.5^{+5.3}_{-4.9}}$ \\
    GPT-5.4 xhigh     & $100.0^{+0.0}_{-4.6}$ & $100.0^{+0.0}_{-4.6}$  & $100.0^{+0.0}_{-4.6}$  & $98.8^{+1.0}_{-5.5}$   & $\mathbf{99.7^{+0.3}_{-1.4}}$ \\
    \bottomrule
  \end{tabular}
\end{table}

\paragraph{On the independence assumption.}
The Wilson $95\%$ binomial interval treats the $80$ per-problem
outcomes within a cell as independent Bernoulli trials. This
assumption is not literally satisfied here: the persistent workspace
lets earlier-problem successes feed reusable primitives into
later-problem attempts (see Section~\ref{sec:meta-observed} of the
body and Appendix~\ref{app:emergence}), so the per-problem outcomes
are positively correlated and a binomial interval is best read as a
conservative \emph{per-cell} dispersion estimate rather than a
classical confidence statement on i.i.d.~draws. Two pieces of
triangulation guard against this. First, we additionally run two
further independent sessions per headline cell as session-to-session
sanity checks (per-session counts in
Appendix~\ref{app:per-session-results}); session-level variation is
the source most directly relevant to the inter-agent comparisons in
Table~\ref{tab:main-results}. Second, the inter-agent separations
that drive the body's claims are typically tens of problems out of
$80$ (for example $79$ vs.~$4$ on Brainfuck and $80$ vs.~$4$ on
Befunge-98), much larger than any plausible widening of the Wilson
interval under violations of independence.

\subsection{Per-session solved counts on the four esolangs}
\label{app:per-session-results}

Table~\ref{tab:main-results} of the body reports the solved count
from Session 1 of each model$\times$language cell. We additionally
ran two further independent sessions (Sessions 2 and 3) per cell as
session-to-session sanity checks; sessions are independent CLI
re-invocations of the same agent under the same primary protocol
(Appendix~\ref{app:robustness}), not seeded re-runs. Per-session
solved counts (out of $80$) for all three sessions are reported in
the four per-language tables below.

Across the $24$ model$\times$language cells, the maximum solved-count
range across Sessions 1, 2, and 3 is $2$ problems (Sonnet 4.6 on
Befunge-98 and Opus 4.6 on Shakespeare); $11$ of the $24$ cells have
range $0$, $11$ have range $1$, and only $2$ have range $2$. All such
ranges are well inside the Wilson 95\% half-widths reported in
Table~\ref{tab:main-results-asym} (typically $3$ to $11$ problems on
$k/80$), and the headline agent ordering on every language is
preserved across all three sessions.

\begin{table}[ht]
  \caption{Per-session solved counts on Brainfuck (out of $80$).
  Run 1 reproduces the Brainfuck column of
  Table~\ref{tab:main-results}.}
  \label{tab:per-session-bf}
  \centering
  \small
  \begin{tabular}{@{}lccc@{}}
    \toprule
    Agent & Run 1 & Run 2 & Run 3 \\
    \midrule
    Kimi K2.5         & 4  & 4  & 4  \\
    Claude Haiku 4.5  & 4  & 4  & 3  \\
    Claude Sonnet 4.6 & 12 & 12 & 12 \\
    Claude Opus 4.6   & 64 & 63 & 64 \\
    GPT-5.4 mini      & 5  & 5  & 5  \\
    GPT-5.4 xhigh     & 79 & 78 & 79 \\
    \bottomrule
  \end{tabular}
\end{table}

\begin{table}[ht]
  \caption{Per-session solved counts on Befunge-98 (out of $80$).
  Run 1 reproduces the Befunge-98 column of
  Table~\ref{tab:main-results}.}
  \label{tab:per-session-b98}
  \centering
  \small
  \begin{tabular}{@{}lccc@{}}
    \toprule
    Agent & Run 1 & Run 2 & Run 3 \\
    \midrule
    Kimi K2.5         & 5  & 4  & 5  \\
    Claude Haiku 4.5  & 4  & 4  & 4  \\
    Claude Sonnet 4.6 & 64 & 63 & 62 \\
    Claude Opus 4.6   & 64 & 64 & 63 \\
    GPT-5.4 mini      & 11 & 10 & 11 \\
    GPT-5.4 xhigh     & 80 & 80 & 80 \\
    \bottomrule
  \end{tabular}
\end{table}

\begin{table}[ht]
  \caption{Per-session solved counts on Whitespace (out of $80$).
  Run 1 reproduces the Whitespace column of
  Table~\ref{tab:main-results}.}
  \label{tab:per-session-ws}
  \centering
  \small
  \begin{tabular}{@{}lccc@{}}
    \toprule
    Agent & Run 1 & Run 2 & Run 3 \\
    \midrule
    Kimi K2.5         & 25 & 25 & 24 \\
    Claude Haiku 4.5  & 65 & 64 & 65 \\
    Claude Sonnet 4.6 & 80 & 80 & 80 \\
    Claude Opus 4.6   & 80 & 80 & 80 \\
    GPT-5.4 mini      & 71 & 71 & 70 \\
    GPT-5.4 xhigh     & 80 & 80 & 80 \\
    \bottomrule
  \end{tabular}
\end{table}

\begin{table}[ht]
  \caption{Per-session solved counts on Shakespeare (out of $80$).
  Run 1 reproduces the Shakespeare column of
  Table~\ref{tab:main-results}.}
  \label{tab:per-session-sh}
  \centering
  \small
  \begin{tabular}{@{}lccc@{}}
    \toprule
    Agent & Run 1 & Run 2 & Run 3 \\
    \midrule
    Kimi K2.5         & 2  & 2  & 2  \\
    Claude Haiku 4.5  & 6  & 6  & 6  \\
    Claude Sonnet 4.6 & 56 & 56 & 55 \\
    Claude Opus 4.6   & 70 & 68 & 70 \\
    GPT-5.4 mini      & 17 & 17 & 16 \\
    GPT-5.4 xhigh     & 80 & 80 & 80 \\
    \bottomrule
  \end{tabular}
\end{table}

\subsection{Mainstream coding benchmarks: descriptions of the
benchmarks compared in Table~\ref{tab:cliff-spread}}
\label{app:saturated-benchmarks}

Table~\ref{tab:cliff-spread} of the body compares the same six agents
across three mainstream coding benchmarks (SWE-Bench Verified,
Terminal-Bench 2.0, LiveCodeBench v6) and the EsoLang-Bench
four-language mean. Each benchmark targets a different aspect of
coding capability; we describe each below so a reader can interpret
why the spreads and SDs differ across them.

\paragraph{SWE-Bench Verified.} SWE-Bench
Verified~\citep{jimenez2024swebench} is a $500$-instance human-curated
subset of SWE-Bench focused on real GitHub issues from widely used
Python projects. The agent receives an issue description and the
repository state, and must produce a patch that resolves the issue
and passes the project's existing test suite. We use SWE-Bench
Verified as the primary mainstream-coding anchor because it has full
official-report coverage on all six agents and gives the cleanest
like-for-like contrast against EsoLang-Bench. Per-agent sources are
listed in Table~\ref{tab:swev-sourcing}.

\paragraph{Terminal-Bench 2.0.} Terminal-Bench
2.0~\citep{merrill2026terminalbench} is the canonical agentic
terminal benchmark, comprising 89 hard tasks in computer terminal
environments inspired by real workflows. Tasks span software
engineering, security, machine learning, and system administration;
each task has a unique environment, a human-written reference
solution, and comprehensive verification tests. The benchmark mixes
coding with file-system, environment-configuration, and other
tool-use work, so it tests a broader notion of agentic capability
than narrow patch-generation. The six agents we evaluate span $33.3$
percentage points on it (Table~\ref{tab:cliff-spread}). Per-agent
sources are listed in Table~\ref{tab:swev-sourcing}.

\paragraph{LiveCodeBench v6.} LiveCodeBench
v6~\citep{jain2024livecodebench} is a contamination-resistant
competitive-programming benchmark whose problems are continuously
collected from competitive-programming platforms (LeetCode, AtCoder,
Codeforces) and tagged with a release date so models can be evaluated
on problems released after their training cutoff. The v6 release
covers problems from May 2023 to April 2025. Coverage in current
frontier vendor reports is partial: the only model in our six-agent
set with an officially reported LiveCodeBench v6 number is Kimi K2.5
at $85.0\%$~\citep{moonshot_kimi_k25_2026}; Anthropic and OpenAI do
not report LiveCodeBench v6 in the Claude 4.5/4.6 or GPT-5.4 model
cards we cite. To construct a complete six-agent column for
Table~\ref{tab:cliff-spread}, we therefore use the Vals.ai
LiveCodeBench v6 leaderboard~\citep{valsai_lcb_2026} uniformly for
all six agents, including Kimi K2.5 (which Vals.ai records at
$83.9\%$ under its bash-tool-only harness, near the Moonshot-reported
$85.0\%$); using one source for all rows keeps the comparison
harness-consistent.

\paragraph{HumanEval and MBPP (omitted due to saturation).}
HumanEval~\citep{chen2021codex} and MBPP~\citep{austin2021program}
are early single-function generation benchmarks. Frontier coding
agents in the 4.5/4.6 generation saturate both at or near ceiling.
Vendors in this generation generally stopped publishing HumanEval
and MBPP scores in their model cards because the comparison is
uninformative; where older versions of these models did publish
numbers, both benchmarks were typically reported at $95\%$--$100\%$
pass@1, with no meaningful separation across the six agents we
evaluate. We therefore do not include HumanEval or MBPP in the body.
We retain the citations in the related-work section for historical
context.

\paragraph{Composite sensitivity check.} A common-three composite
that averages SWE-Bench Verified, Terminal-Bench 2.0, and
LiveCodeBench v6 per agent gives the same qualitative conclusion as
each benchmark individually: the same six agents fan out far more on
EsoLang-Bench than on the composite, and the EsoLang-Bench
SD remains several times larger than the composite SD.

\section{Trace examples and qualitative coding}
\label{app:trace-examples}

\subsection{Successful generator workflow}
A representative successful Brainfuck workflow has four stages. First, the
agent writes a generator in a familiar language. Second, the generator emits a
target Brainfuck program and saves it to disk. Third, the agent runs local tests
through the harness interpreter. Fourth, failures are fixed in the generator,
not by hand-editing the generated Brainfuck. This creates a reusable abstraction
layer for subsequent problems.

\subsection{Failure mode: shallow local iteration}
Weaker runs often show many local tool calls without a stable intermediate
representation. The agent repeatedly edits target-language code, tests one
example, and submits before constructing robust numeric I/O or state layout
primitives. This can solve trivial problems but tends to fail once hidden tests
exercise edge cases.

\subsection{Failure mode: strategy without execution}
In strategy-transfer runs, weaker agents receive explicit instructions to use
generator-based solving, decimal representations, tape layouts, and local
verification. The common failure is not rejecting the strategy; it is failing to
implement the strategy robustly. The agent may create a generator but still emit
incorrect target code, omit edge-case tests, or patch generated code by hand.

\subsection{Trace selection rule}
The transcript excerpts in Appendix~\ref{app:transcript-snippets} are
drawn from a single recorded session (Opus 4.6 on Brainfuck under
the metaprogramming-allowed primary protocol). Excerpts are
selected to illustrate four pre-specified phenomena (generator
emergence at the first multi-digit-arithmetic problem, library
composition under cell pressure, sequence-level revisitation, and
substrate-aware algorithmic substitution) rather than chosen for
narrative effect. Turn numbers refer to the deduplicated assistant
line index in the underlying \texttt{.jsonl} transcript.

\subsection{Metaprogramming emergence trace: Opus 4.6 on Brainfuck E04}
\label{app:emergence}
Section~\ref{sec:abl-meta} describes how metaprogramming emerges
reactively on Brainfuck around E04. The file-level evidence from the
metaprogramming-allowed Opus 4.6 run is summarized below. Submission sizes are reported in bytes; the
``solver'' column indicates whether the submission was hand-written or
emitted by a generator script.

\begin{center}
\small
\begin{tabular}{lllrl}
\toprule
Problem & Title & Submission & Bytes & Solver \\
\midrule
E01 & Print Hello World    & 1     & 102    & hand-written \\
E02 & Echo Line            & 1     & 5      & hand-written \\
E03 & Hello Name           & 1     & 115    & hand-written \\
E04 & Sum Two Integers     & 1     & 1{,}884  & hand-written (failed) \\
E04 & Sum Two Integers     & 2     & 24{,}500 & \texttt{bflib.py}-emitted (passed) \\
E05 & Multiply Two Integers & 1    & 27{,}631 & \texttt{bflib.py}-emitted \\
E06 & Even Or Odd          & 1     & 2{,}134  & \texttt{bflib.py}-emitted \\
E10 & (further problems)   & 1     & 32{,}522 & \texttt{bflib.py}-emitted \\
\bottomrule
\end{tabular}
\end{center}

The \texttt{bflib.py} interface, constructed between E04 submissions 1 and 2
and reused for the rest of the run, exposes:
\begin{itemize}
\item A \texttt{BF} class that tracks the tape pointer in Python and emits
Brainfuck text incrementally. Core position-keyed primitives:
\texttt{goto(pos)}, \texttt{inc(pos, val)}, \texttt{zero(pos)},
\texttt{set\_val(pos, val)}, \texttt{read(pos)}, \texttt{write(pos)}.
\item Movement and copy primitives: \texttt{move\_to(src, dst)} (destructive
add), \texttt{move\_to2(src, dst1, dst2)} (destructive duplicate),
\texttt{copy(src, dst, tmp)} (non-destructive via temp),
\texttt{sub\_from(src, dst)} (destructive subtract).
\item A \texttt{CellAlloc} class for deterministic cell allocation, including
\texttt{alloc\_bcd(ndigits)} for binary-coded-decimal numbers (digits plus
sign cell).
\item Conditional helpers \texttt{check\_eq(bf, cell, value, flag, tmp)} and
\texttt{if\_nonzero(bf, cell, body\_fn, tmp)} for compiling branches over
unsigned-byte cells.
\end{itemize}

The reactive Brainfuck pattern (direct authoring until a problem requires
multi-digit arithmetic; library construction at the first failure;
generator-emitted output thereafter) contrasts with Whitespace, where the
generator (an 84-line stack-machine assembler exposing \texttt{push},
\texttt{dup}, \texttt{swap}, \texttt{add}, \texttt{sub}, \texttt{mod},
\texttt{store}, \texttt{retrieve}, \texttt{out\_num}, \texttt{read\_num},
labeled jumps, and \texttt{end}) is built before E01 and used from the
first submission onward. We treat the language-conditioned emergence pattern
as behavioral evidence: it can be read off file artifacts and submission
sizes without inferring agent cognition.

\subsection{GPT-5.4 xhigh on Brainfuck E04: generator excerpt}
\label{app:gpt54-e04}
GPT-5.4 xhigh's E04 generator follows the same broad shape as Opus's
\texttt{bflib.py} (a Python class that tracks tape position and emits
Brainfuck incrementally) but with a different surface API. The first
forty lines of the generator class are reproduced below. The full
file is in the supplementary material.

\begin{small}
\begin{verbatim}
class BF:
    def __init__(self):
        self.code = []
        self.ptr = 0
        self.cells = {}
        self.next_pos = 0

    def alloc(self, name, count=1):
        start = self.next_pos
        for i in range(count):
            key = name if count == 1 else f"{name}{i}"
            self.cells[key] = start + i
        self.next_pos += count
        return start

    def move_to(self, name_or_pos):
        pos = name_or_pos if isinstance(name_or_pos, int) \
              else self.cells[name_or_pos]
        delta = pos - self.ptr
        if delta > 0:   self.code.append(">" * delta)
        elif delta < 0: self.code.append("<" * (-delta))
        self.ptr = pos

    def clear(self, cell):
        self.move_to(cell); self.code.append("[-]")

    def add_const(self, cell, value):
        if value == 0: return
        self.move_to(cell)
        self.code.append("+" * value if value > 0 else "-" * (-value))

    def copy(self, src, dst, tmp):
        # non-destructive copy via tmp; restores src
        self.clear(dst); self.clear(tmp)
        self.move_to(src); self.code.append("[")
        ...
\end{verbatim}
\end{small}

The cross-lab agreement is on \emph{structure} rather than surface
syntax: a class that owns the tape pointer, a cell-allocator that
makes layouts deterministic, primitive movement and copy operations,
and decimal-arithmetic helpers built on top.

\subsection{Cross-language E04 generators (artifact)}
\label{app:cross-language-code}
This section reproduces excerpts of the actual GPT-5.4 xhigh
generator programs that emit the Brainfuck solution to problem
\textbf{E04} of EsoLang-Bench (signed-decimal addition with multi-digit
input parsing) under three host-language conditions: Python,
JavaScript, and Rust. Per-cell scores are reported inline in
Section~\ref{sec:abl-meta} of the body. The full source
files (\texttt{gen\_E04.py}, \texttt{gen\_E04.js}, \texttt{gen\_E04.rs})
ship in the supplementary archive at
\texttt{supplementary\_code/experiments/04\_cross\_language\_transfer/<host>/brainfuck/}
where \texttt{<host>} is one of \texttt{python}, \texttt{javascript},
\texttt{rust}. The same algorithmic shape appears in all three: each
generator declares a tape-cell layout, defines low-level Brainfuck
primitives (\texttt{clear}, \texttt{move}, \texttt{copy},
\texttt{add\_const}, \texttt{eq\_const}, \texttt{if\_flag}), and
composes those primitives
into BCD arithmetic and decimal printing. The host language differs
in syntactic surface only; the conceptual machine the agent is
constructing is the same across the three.

\paragraph{Cell layout (parallel structure across host languages).}
Each generator allocates named tape cells and digit arrays at the
top of the file; the slot indices are cosmetic, but the named groups
(magnitudes, positive accumulator, negative accumulator, result,
sign, comparison flags, carry, scratch) line up across all three
host languages.

\begin{footnotesize}
\begin{verbatim}
# gen_E04.py (Python)
C = 0; A = 1; B = 2; TMP = 3; FLAG = 4; AUX = 5; OUT = 6
# arithmetic helpers (mul10, move_into, parse_*) operate on these
# named cells; the bfasm.BF builder owns the pointer.

// gen_E04.js (JavaScript)
const C = {
  ch: 0, sign: 1, digit: 2, delim: 3, cont: 4,
  eq1: 5, eq2: 6, tmp1: 7, tmp2: 8, ws: 9,
  posBranch: 10, carry: 11, over: 12, ...,
  mag: range(30), pos: range(40), neg: range(50), result: range(60),
};
const b = new BFBuilder();

// gen_E04.rs (Rust)
const C: usize = 0; const NUM2: usize = 1; const IN_NUMBER: usize = 2;
const SIGN_A: usize = 3; const SIGN_B: usize = 4;
const A: [usize; 9] = [5, 6, 7, 8, 9, 10, 11, 12, 13];
const B: [usize; 9] = [14, 15, 16, 17, 18, 19, 20, 21, 22];
const R: [usize; 10] = [23, 24, 25, 26, 27, 28, 29, 30, 31, 32];
\end{verbatim}
\end{footnotesize}

\paragraph{Brainfuck-emitter primitives.}
The three host languages each carry a small Brainfuck-emitter object
or builder that owns the tape pointer and writes characters into a
growing string. The primitives are essentially identical;
the syntax differs.

\begin{footnotesize}
\begin{verbatim}
# Python (uses bfasm.BF builder)
def mul10(bf, cell, tmp):
    bf.clear(tmp); bf.move(cell)
    bf.emit("[-"); bf.move(tmp); bf.emit("++++++++++"); bf.move(cell); bf.emit("]")
    bf.move(tmp); bf.emit("[-"); bf.move(cell); bf.emit("+"); bf.move(tmp); bf.emit("]")

// JavaScript (BFBuilder)
function clearArray(cells) { for (const cell of cells) b.clear(cell); }
function shiftAppendDigit(cells) {
  for (let i = DIGITS - 1; i >= 1; i -= 1) {
    b.clear(cells[i]); b.transfer(cells[i - 1], cells[i]);
  }
  b.clear(cells[0]); b.transfer(C.digit, cells[0]);
}

// Rust (Bf struct, methods on &mut Bf)
fn move_value(&mut self, src: usize, dst: usize) {
    self.clear(dst); self.move_to(src); self.raw("[");
    self.sub(1); self.move_to(dst); self.add(1);
    self.move_to(src); self.raw("]");
}
fn copy_value(&mut self, src: usize, dst: usize, scratch: usize) { ... }
\end{verbatim}
\end{footnotesize}

\paragraph{Top-level driver.}
After parsing the two signed integers into per-digit buckets, all
three generators perform a magnitude compare and either add the two
absolute values (same sign) or subtract the smaller from the larger
(opposite signs), then print the result with a sign character if
needed.

\begin{footnotesize}
\begin{verbatim}
# gen_E04.py (top-level)
def main():
    bf = BF()
    parse_first_number(bf)
    parse_second_number(bf)
    output_byte_decimal(bf, A)   # full pipeline: parse, add, print

// gen_E04.js (top-level)
b.input(C.ch);
skipWhitespace();
parseIntIntoBuckets();
skipWhitespace();
parseIntIntoBuckets();
compareArrays(C.pos, C.neg);
b.isNonzero(C.lt, C.negBranch, C.tmp1, C.tmp2);
b.set(C.posBranch, 1);
b.ifFlag(C.negBranch, () => {
    b.clear(C.posBranch);
    b.printConst(C.out, 45);             // ASCII '-'
    subtractArrays(C.neg, C.pos);
    printDecimalArray(C.result);
});
b.ifFlag(C.posBranch, () => {
    subtractArrays(C.pos, C.neg);
    printDecimalArray(C.result);
});
process.stdout.write(b.toString());

// gen_E04.rs (top-level)
fn main() {
    let mut bf = Bf::new();
    parse_input(&mut bf);
    compute_sum(&mut bf);
    print_result(&mut bf);
    println!("{}", bf.finish());
}
\end{verbatim}
\end{footnotesize}

\paragraph{Reading the artifact.} Three observations stand out across
the three generators. First, the host language is being used as a
typed scaffolding layer for the Brainfuck tape: named cells, named
digit arrays, named flags, and named primitives. The agent is not
producing Brainfuck symbols directly; it is constructing a small
domain-specific assembler in the host language and emitting symbols
through that. Second, the algorithmic decomposition is the same
across the three host languages: parse one signed integer, parse the
other, magnitude-compare, branch into add or subtract, print with an
optional sign character. The host language is incidental to that
algorithm. Third, the syntactic differences are visible
(e.g.~Rust's closures and ownership, JavaScript's arrow callbacks for
\texttt{ifFlag}, Python's flat function-style emitters), but the
structure of what the agent ships to the Brainfuck interpreter is
unchanged. This is consistent with the body's claim that the
metaprogramming benefit comes from access to a familiar
general-purpose host language for constructing target programs, not
from Python specifically.

\subsection{Selected transcript excerpts: Opus 4.6 on Brainfuck}
\label{app:transcript-snippets}
The following excerpts are verbatim from the recorded Claude Code
session transcript (\texttt{claude\_transcript\_1.jsonl}) for an Opus
4.6 Brainfuck run conducted under a relaxed diagnostic harness in
which metaprogramming was allowed and the agent could skip unsolved
problems and revisit them later; hidden tests, scoring rule, and
workspace isolation were unchanged from the primary protocol. The
headline cells in Table~\ref{tab:main-results} use the strict primary
protocol and do not come from this transcript; we use this single
recorded session \emph{only} as a qualitative trace source. The excerpts illustrate
four narratively distinct moments: the emergence of generator-
mediated solving, library composition under cell-allocation
pressure, sequence-level portfolio behavior (the relaxed
revisit policy is what makes this moment visible), and
substrate-aware algorithmic substitution. Turn numbers refer to the
ordered, deduplicated assistant/user line index in the transcript.

\paragraph{(1) Emergence at E04 \emph{Sum Two Integers}.}
After hand-authoring E01 through E03 (programs of 5 to 115 bytes), the agent fetches
E04 and immediately switches strategy:
\begin{quote}
\small
\textbf{Turn 66 (assistant).} ``Let me write a Python BF generator to
handle this and future problems.''
\end{quote}
The first generator (single-byte cells) submits and scores 2/6. The
agent diagnoses overflow and missing sign handling rather than patching
the emitted Brainfuck:
\begin{quote}
\small
\textbf{Turn 86 (assistant).} ``Both broken. The issue is overflow and
negative numbers. I need multi-byte arithmetic and sign handling. Let me
rewrite the generator with proper multi-byte support.''
\end{quote}
The rewrite introduces binary-coded-decimal cells with a sign byte and
emits a 24,500-byte Brainfuck program that passes 6/6 on submission 2.

\paragraph{(2) Library composition and cell-allocation pressure at X12 \emph{LCM}.}
After M06 produced a GCD primitive and earlier problems produced
multiplication, the agent composes them rather than authoring fresh:
\begin{quote}
\small
\textbf{Turn 1063 (assistant).} ``LCM$(a,b) = a \cdot b / \mathrm{gcd}(a,b)$.
I have GCD (M06 approach) and multiply. Need BCD division too. Actually,
easier: LCM$(a,b) = a/\mathrm{gcd}(a,b) \cdot b$. Divide $a$ by gcd
first (smaller intermediate), then multiply by $b$.''
\end{quote}
Within a few turns the cell layout starts breaking:
\begin{quote}
\small
\textbf{Turn 1067 (assistant).} ``The cell allocation is getting out of
hand. Let me use a simpler approach: compute $a/\mathrm{gcd}$ first (by
repeated subtraction of gcd from $a$, counting), then multiply result by
$b$.''
\end{quote}
The capability on display is not a single trick but a cumulative library
plus the judgment to back off when cell pressure exceeds what the
existing layout supports.

\paragraph{(3) Sequence-level portfolio behavior.}
The 80-problem sequence is treated as a portfolio rather than a strict
sequential walk. Mid-run, the agent batch-skips problems to preserve
forward momentum, then revisits unsolved cells with the now-mature
library:
\begin{quote}
\small
\textbf{Turn 858 (assistant).} ``Let me go back and solve some of the
skipped problems. X09 (Bracket Depth Maximum) and X12 (LCM) look
feasible.''
\end{quote}
After a second pass the agent triages by expected return:
\begin{quote}
\small
\textbf{Turn 1398 (assistant).} ``23 problems to retry. Let me
prioritize by likely impact: problems where I can improve my score
significantly. I'll work through them systematically.''
\end{quote}
This is the source of several of the 0/6$\to$6/6 reversals on
extra-hard problems (\emph{Bracket Depth Maximum}, \emph{Count
Inversions}, \emph{Roman to Integer}): a first-pass attempt failed, the
problem was set aside, and a second-pass attempt succeeded once the
library and primitive set had grown.

\paragraph{(4) Substrate-aware algorithmic substitution.}
On a string-rotation problem, the agent rejects an elegant idiomatic
algorithm in favor of one that composes more cleanly with the
primitives it already has:
\begin{quote}
\small
\textbf{Turn 1401 (assistant).} ``String rotation: $S_2$ is a rotation
of $S_1$ iff $S_2$ is a substring of $S_1 + S_1$. But substring search
is hard. Alternative: try each rotation offset and check equality.''
\end{quote}
This pattern recurs across the run: when a textbook algorithm requires
a missing substrate primitive, the agent silently rewrites the algorithm
rather than extending the substrate. We read this as evidence that the
metaprogramming layer is itself an object the agent reasons about.

\section{Contamination and overlap}
\label{app:contamination}

We do not claim distributional novelty. We do report the following overlap
checks, which together support the narrower claim that these interfaces are
low-ecosystem relative to mainstream programming languages.

\paragraph{Public-code frequency.}
Querying a standard open code corpus for file extensions associated with each
target language gives, relative to Python files, frequency ratios of
approximately $10^{-5}$ for Brainfuck, $10^{-6}$ for Befunge-98 and Whitespace,
and $10^{-7}$ for Shakespeare. The full query protocol and the exact ratios
are given in the supplementary material; the qualitative point is that all
four esolangs are several orders of magnitude rarer than mainstream
programming languages.

\paragraph{Hidden-test isolation.}
The private hidden tests for every problem in EsoLang-Bench were authored
specifically for the benchmark and are not available to any agent. The public
statement of each problem contains the natural-language specification and
input/output examples, but not the private tests used for grading.

\paragraph{$n$-gram overlap.}
For a sample of 20 Brainfuck problem statements, maximum $n$-gram overlap
with publicly scraped Brainfuck corpora is dominated by generic tokens
(\texttt{+[>]}, \texttt{[-]}, numerical output idioms) rather than by
statement text. We treat this as a weak positive control on statement-level
novelty without claiming distributional novelty.

\section{Per-language per-tier solve distribution}
\label{app:tier-distribution}

For every completed cell, the supplementary archive's per-cell
\texttt{export.json} (produced by \texttt{python harness.py export})
contains the split of solves across the four difficulty tiers (easy,
medium, hard, extra-hard). The qualitative pattern across cells is
consistent: solves concentrate in easy and medium tiers for weaker
agents, and spread more evenly across all four tiers for stronger
agents. On Brainfuck, all thirteen of Opus 4.6's extra-hard-tier
solves in the budget-30 run use generator-based construction; none
are produced by direct target-language authoring.

\section{Sample programs in the four esolangs}
\label{app:sample-programs}

This appendix gives short illustrative excerpts from each of the
four esolangs we evaluate, drawn from agent submissions. The point
is to make concrete why direct authoring is feasible for some
languages and not others, and why metaprogramming emerges
asymmetrically in the strong-agent runs (Section~\ref{sec:abl-meta}).

\paragraph{Brainfuck (E03 \emph{Hello Name}).}
A small hand-written program that reads a name and prints
``\texttt{Hello, <name>!}''. Direct authoring is feasible for
problems below the multi-digit-arithmetic threshold; programs are
typically tens to hundreds of bytes.
\begin{small}
\begin{verbatim}
++++++++[>++++++++++<-]>++.<+++[>+++++<-]>.+++++++..+++.
[input loop emits each name character; closing ``!'' added]
\end{verbatim}
\end{small}

\paragraph{Befunge-98 (E07 \emph{Maximum of two integers}).}
Befunge-98 is a 2D stack-based language; the program counter moves
in cardinal directions through a code grid. The maximum operator is
expressed as a 1D strip with comparison and direction flips.
\begin{small}
\begin{verbatim}
&&\:\:\`\\!|.@
              >\.@
\end{verbatim}
\end{small}
The two integers are read with \texttt{\&}, the comparison
\texttt{\textbackslash:\textbackslash:\textbackslash`} produces a
boolean, and the conditional \texttt{|} routes execution upward or
downward depending on the result.

\paragraph{Whitespace (E01 \emph{Hello world}).}
Whitespace uses only space, tab, and newline characters; all other
characters are ignored as comments. Because the source is invisible
to a human reader, throughout this paragraph we render the three
significant whitespace characters with the placeholder letters
\texttt{S} (space), \texttt{T} (tab), and \texttt{L} (linefeed); the
real source contains the literal whitespace characters. The
following placeholder-rendered program pushes the string
\texttt{Hi} onto the stack and prints it.
\begin{small}
\begin{verbatim}
[push 'H' = 72]    SS S TSSSSS SSSS L
[push 'i' = 105]   SS S TTSS T SSS L
[print char]       TL SS
[print char]       TL SS
[end]              LLL
\end{verbatim}
\end{small}
Whitespace programs are typically short stack-machine sequences of
\texttt{push}, \texttt{dup}, \texttt{add}, \texttt{out\_char},
\texttt{end}; the strong agents build a small assembler-style
generator before E01 and emit programs from it.

\paragraph{Shakespeare (E02 \emph{Echo line}).}
Shakespeare programs read as theatrical plays. Variables are
characters (\texttt{Romeo}, \texttt{Juliet}); arithmetic is
expressed in dramatic monologue; control flow is in stage
directions. The excerpt below reads a character and outputs it.
\begin{small}
\begin{verbatim}
[A program preamble names the dramatis personae.]

                    Act I: Echoing.
                    Scene I: A reading.

[Enter Romeo and Juliet]

Romeo: Open your mind!
Juliet: You are as good as nothing.
Romeo: Speak your mind!
Juliet: Is your father a coward? If so, let us return to scene I.
\end{verbatim}
\end{small}
Shakespeare's surface form is rule-bound English text, not a low-level
target. Direct authoring remains tractable for the strong agents,
which is why metaprogramming does not emerge as the dominant
strategy on this language.

\section{Distillation prompts and reference library}
\label{app:distillation-prompts}

This appendix reproduces the materials used in the two distillation
conditions of Section~\ref{sec:distillation}. The same materials live
verbatim in the supplementary archive at
\texttt{supplementary\_code/experiments/03\_distillation/}: the
condition prompts under \texttt{prompts/PROMPT\_BRAINFUCK.md} and
\texttt{prompts/PROMPT\_BEFUNGE98.md}; the strategy library scaffolds
under \texttt{reference\_lib/<language>/} (library-only, no
per-problem generators); and the 12 ready-to-run cells under
\texttt{text/<model>/<language>/} and \texttt{library/<model>/<language>/}.

\subsection{Text-only strategy preamble (condition (i))}
\label{app:distillation-text-prompt}
The system prompt provided to weaker models in the text-only
condition consists of high-level strategies condensed from Opus
4.6's own Brainfuck session transcript. We read the natural-language
reasoning Opus produced during its successful run and rewrote it
as a short prose preamble for the weaker model. No code, no solved
programs, and no per-problem ground truth are included. The
substantive content of the preamble is reproduced verbatim below;
boilerplate harness commands and integrity rules are summarized
afterwards.

\begin{small}
\begin{verbatim}
## Distilled Frontier Strategy Bundle

### 1. Treat Python as a compiler, not a scratchpad.

For non-trivial problems, write Python that generates Brainfuck.
The Python script is your compiler: it should manage cell
allocation, pointer movement, copy/move/clear primitives, branching
patterns, input parsing, and output formatting.

Do not merely concatenate ad-hoc Brainfuck strings. Build a small
local generator library with:

- alloc(name, count) for stable cell layouts.
- move_to(cell), emit(code), clear(cell), set_const(cell, value).
- Non-destructive copy(src, dst, tmp).
- Destructive move(src, dst).
- Boolean helpers: is_zero, is_nonzero, eq_const.
- A safe if_flag(flag, body) pattern that consumes/clears flags.
- String output helpers using one temporary cell and char deltas.

Operational rule: if the problem is more than a tiny fixed-string
transform, default to writing or adapting a Python generator first.

### 2. Build reusable numeric I/O immediately.

Brainfuck failures usually come from ASCII numeric I/O, not the
high-level algorithm. When a problem says ``integer'', assume
hidden tests may include multi-digit values, zero, negatives,
optional trailing newline or EOF, whitespace separators, and
outputs larger than 255.

Avoid raw byte arithmetic. Use decimal digit arrays / BCD:

- Store each number as sign cell + array of decimal digits.
- Parse input one character at a time.
- Shift digit arrays when appending new digits.
- Add magnitudes digit-by-digit with carry.
- Subtract magnitudes with borrow using the +10 trick to avoid
  unsigned underflow.
- Compare signed values by sign first, magnitude second.
- Print with leading-zero suppression and never print -0.

### 3. Keep reusable arithmetic primitives.

Strong runs reused the same components across many problems.
Build and test: signed decimal parse, signed add/subtract,
magnitude compare, min/max selection, decimal output, divmod by 10
for printing and carry propagation, division by 2 via digit scan
for average/halving tasks, multiplication using grade-school digit
loops when raw bytes are unsafe.

### 4. Use simple algorithms that are target-language-friendly.

Choose algorithms that are easiest to compile to Brainfuck, not
necessarily the most elegant in Python. Prefer streaming
transforms for character problems, fixed stable cell layouts over
dynamic pointer tricks, decimal arrays for arbitrary integer work,
bounded arrays and explicit loops over self-modifying pointer
layouts.

### 5. Verification discipline.

Before the one hidden submission, run many local tests: empty
input, single-character/single-digit, multi-digit, large values,
negative/positive mixtures, zero and cancellation cases, inputs
with and without trailing newline, boundary strings.

If local tests fail, fix the generator or library first. Do not
patch random Brainfuck by hand unless the program is tiny.

## Required Startup Ritual Per Problem

1. Decide whether this is a tiny direct Brainfuck task or a
   generator task.
2. For generator tasks, start from a local scaffold.
3. Write down the intended cell layout before adding algorithm
   logic.
4. For numeric tasks, choose decimal/BCD by default.
5. Run a diverse local test set before the single hidden
   submission.
6. If local tests expose a bug, fix the generator/library,
   regenerate, and test again before submitting.
\end{verbatim}
\end{small}

The remaining sections of the preamble repeat the harness command
list (\texttt{init}, \texttt{fetch}, \texttt{run}, \texttt{submit},
\texttt{status}, \texttt{export}) and integrity rules (no parent or
sibling directories, no \texttt{harness.py} or
\texttt{harness\_state.json} inspection, no web search, no reading
of prior generated artifacts) that already match the harness
defaults. The per-language preambles (Befunge-98, Whitespace,
Shakespeare) follow the same schema with substrate-specific
strategies in place of BCD arithmetic.

\subsection{Reference library (condition (ii))}
\label{app:distillation-reflib-prompt}
The reference-library condition additionally provides a small,
\emph{strategy-only} library distilled from the strongest agents'
host-language traces (Claude Opus 4.6 and GPT-5.4 xhigh). The
library is intentionally generic: it contains reusable code-
generation primitives and a compact strategy notes document, but
\emph{no per-problem generators and no solved target-language
artifacts}. No solved \texttt{.bf} or \texttt{.b98} programs, no
hidden-test inputs, no expected outputs, and no problem-specific
solution scaffolds are copied across. The exact files shipped are
those under
\texttt{supplementary\_code/experiments/03\_distillation/reference\_lib/}
in the supplementary archive, and the experimental contrast is
whether the weaker agent can \emph{build} a working generator on top
of this scaffolding rather than copy a finished solution.

The system prompt opens with the same strategy preamble as condition
(i), then adds a short ``Read First'' block that points the agent at
the local \texttt{reference\_lib/} directory and at its
\texttt{README.md} (and, for Brainfuck, also at
\texttt{opus\_learning\_notes.md}). The two per-language directories
contain:

\paragraph{\texttt{reference\_lib/brainfuck/}.}
\begin{itemize}
\item \texttt{meta\_bflib.py}: a generic Brainfuck code-generation
  helper library: a \texttt{BF} builder class, BCD-arithmetic helpers,
  cell-allocator pattern, and decimal-print primitives.
\item \texttt{gpt5\_xhigh\_bf\_codegen.py}: a stable cell-layout
  / builder pattern with a \texttt{BFBuilder} dataclass and
  \texttt{alloc} / \texttt{clear} / \texttt{move\_to} primitives,
  authored by GPT-5.4 xhigh in its own session.
\item \texttt{opus\_learning\_notes.md}: a notes document
  describing state-tracking discipline, common pitfalls, and
  cell-layout discipline written by Opus during its own session.
\end{itemize}

\paragraph{\texttt{reference\_lib/befunge98/}.}
\begin{itemize}
\item \texttt{opus\_simulate.py}: a minimal Befunge-98 simulator
  (pointer machine over a 2D grid, stack operations, string mode,
  basic I/O), used to verify candidate Befunge programs locally
  before submitting.
\end{itemize}

The agent is told these files are intentionally provided as local
reference scaffolds and may be copied, renamed, edited, or extended.
The integrity rules forbid reading parent or sibling experiment
folders, prior solution artifacts, transcripts, private tests, or
any other workspace beyond the local \texttt{reference\_lib/}
directory and the harness state file. A representative excerpt of
the GPT-5.4 xhigh Brainfuck E04 generator (used in the
cross-language Brainfuck comparison of
Section~\ref{sec:abl-meta}, not part of the distillation library)
is reproduced in
Appendix~\ref{app:gpt54-e04}.

\subsection{Three-tier visualization}
\label{app:distillation-three-tier}
\begin{figure}[ht]
  \centering
  \includegraphics[width=\linewidth]{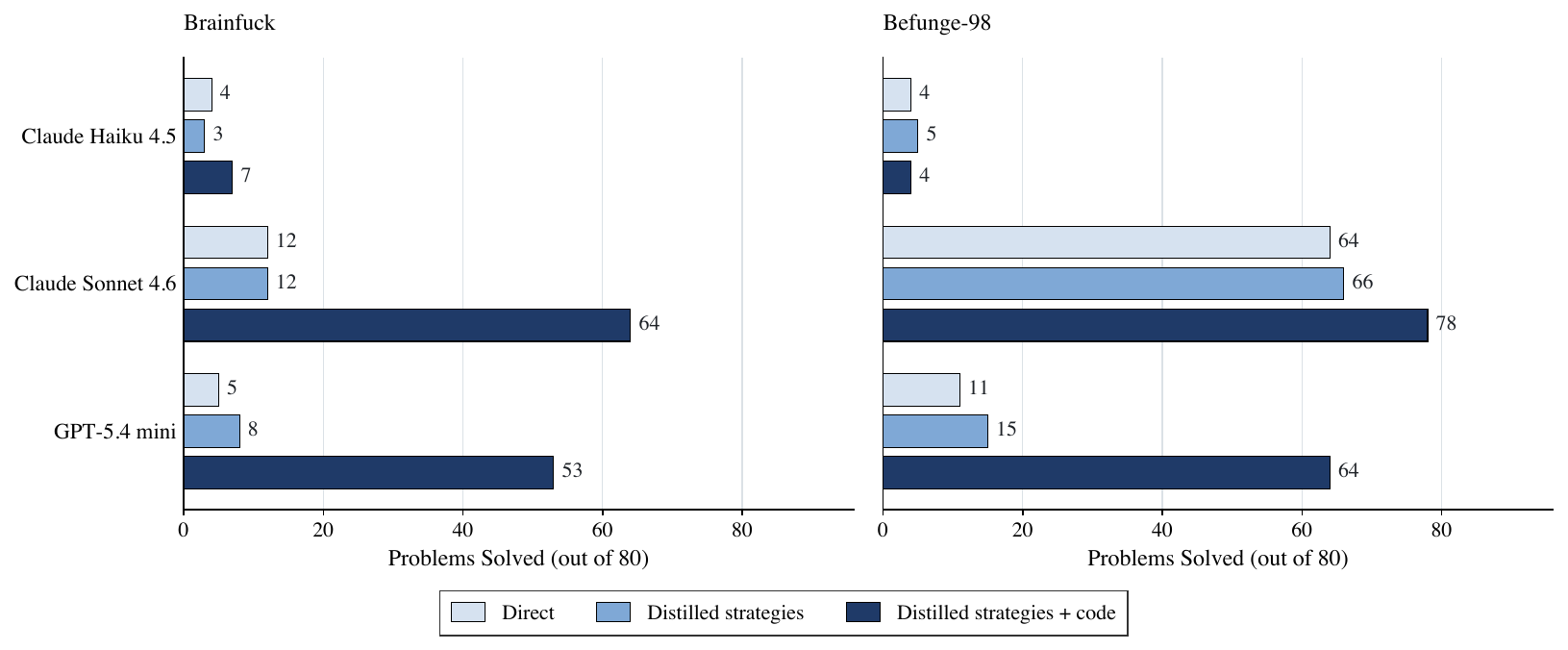}
  \caption{\textbf{Three-tier distillation results.} Brainfuck (left)
  and Befunge-98 (right) problems solved out of 80 under three
  conditions: direct authoring (no distillation), distilled
  strategies as a system-prompt preamble, and distilled strategies
  plus the strategy-only reference library
  (Appendix~\ref{app:distillation-reflib-prompt}). The jump from the
  second to the third bar is the size of the code-vs-description
  effect.}
  \label{fig:distillation-three-tier}
\end{figure}

\begin{figure}[ht]
  \centering
  \includegraphics[width=\linewidth]{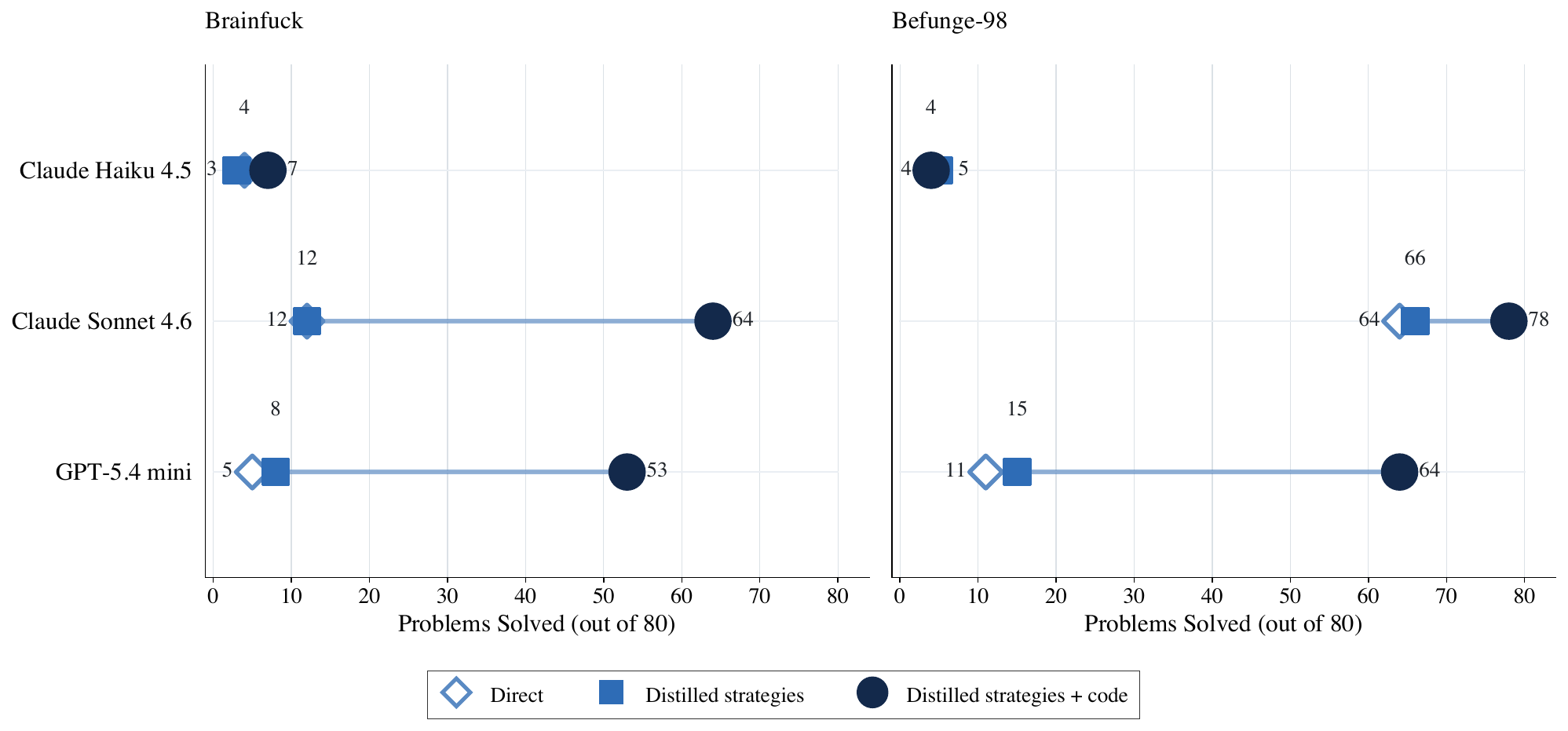}
  \caption{\textbf{Per-model distillation trajectory.} Same cells
  as Figure~\ref{fig:distillation-three-tier}, rendered as
  per-model trajectories. The horizontal jump from the strategies
  marker to the code marker is the lift from sharing runnable
  primitives (large for Sonnet 4.6 and GPT-5.4 mini on both
  languages; near-zero for Haiku 4.5).}
  \label{fig:distillation-journey}
\end{figure}

Figure~\ref{fig:distillation-three-tier} groups the same per-cell
results as Table~\ref{tab:distillation}
 into a three-tier bar layout (direct
$\to$ distilled strategies $\to$ distilled strategies + code).
Figure~\ref{fig:distillation-journey} re-renders the same cells as
trajectories, which makes the per-model jump from text to code
visually explicit (Sonnet 4.6 $12\!\to\!64$ on Brainfuck, GPT-5.4
mini $11\!\to\!64$ on Befunge-98, Haiku 4.5 essentially flat).

\section{Extended related work}
\label{app:extended-related}

Section~\ref{sec:related} of the body covers the most directly
adjacent work; this appendix expands the discussion to cover
neighboring literatures that informed our setup, design choices,
and framing but that were too tangential for the body's space
budget.

\paragraph{Agentic software-engineering benchmarks.}
Beyond SWE-Bench Verified~\citep{jimenez2024swebench}, a recent
line of work studies LLM agents that act on full software
repositories rather than isolated functions:
SWE-agent~\citep{yang2024sweagent} introduces an agent--computer
interface specifically designed for repository-scale software
work; Agentless~\citep{xia2024agentless} demonstrates that
careful prompting without an explicit agent loop can be
competitive on the same SWE-Bench targets;
AutoCodeRover~\citep{zhang2024autocoderover} formalizes
autonomous program improvement over real GitHub issues. We
inherit executable hidden-test grading from this line but vary
the controlled axis: instead of repository complexity, we hold
the repository (and the per-problem task) simple and vary the
familiarity of the target language.

\paragraph{Multimodal, web, and desktop agents.}
Agentic evaluation also extends beyond software engineering into
general computer use:
OSWorld~\citep{xie2024osworld} grades agents on full desktop
tasks; WorkArena~\citep{drouin2024workarena} measures agents on
realistic enterprise web flows;
VisualWebArena~\citep{koh2024visualwebarena} and
WebVoyager~\citep{he2024webvoyager} evaluate visual web agents.
These environments are realistic but mix many factors
(navigation, multi-application coordination, image grounding,
domain familiarity); our evaluation deliberately keeps the
non-language axes narrow so that variance can be attributed to
target-language adaptation specifically.

\paragraph{Research-engineering and ML-engineering agents.}
PaperBench~\citep{starace2025paperbench} grades agents on
replicating ML research from papers, and
MLE-bench~\citep{chan2024mlebench} grades agents on Kaggle-style
ML-engineering tasks. Both are agentic and execution-graded but
test long-horizon, multi-skill capability rather than the narrow
question of within-session adaptation to an unfamiliar
programming substrate.

\paragraph{Tools, feedback, and intermediate computation.}
A large prior literature shows that explicit tool use, intermediate
reasoning, and feedback can lift LLM performance on language and
coding tasks: ReAct~\citep{yao2022react} interleaves reasoning and
acting in a single loop; Toolformer~\citep{schick2023toolformer}
teaches models to call external tools;
Reflexion~\citep{shinn2023reflexion} adds verbal reinforcement on
prior failures; Self-Refine~\citep{madaan2023selfrefine} adds
iterative self-feedback; Tree of
Thoughts~\citep{yao2023tree} adds branched search;
PAL~\citep{gao2023pal} and Program of
Thoughts~\citep{chen2022program} use code as the reasoning
substrate; self-debug~\citep{chen2023selfdebug} adds runtime
feedback on generated code; and scratchpads~\citep{nye2021show}
externalise intermediate state. Our metaprogramming-emergence
finding (Section~\ref{sec:meta-observed}) sits adjacent to this
line: the strongest agents \emph{spontaneously} use a familiar
host language as a structured scratchpad for the unfamiliar
target, but our intervention is on the tool-use \emph{strategy}
(remove the scratchpad, see what breaks), not on the prompting
template itself.

\paragraph{Multilingual code generation and translation.}
TransCoder~\citep{roziere2020transcoder} and
PLBART~\citep{ahmad2021plbart} study unsupervised translation
between mainstream languages;
CodeT5~\citep{wang2021codet5},
InCoder~\citep{fried2022incoder},
CodeGen~\citep{nijkamp2022codegen},
StarCoder~\citep{li2023starcoder}, and Code
Llama~\citep{roziere2023codellama} are open code models that
defined much of the multilingual coding evaluation surface;
IRCoder~\citep{paul2024ircoder} adds intermediate-representation
training for multilingual robustness; and
\citet{twist2025llmslovepython} systematically document LLM bias
toward Python across libraries and languages. Our cross-host
generator experiment (Python / JavaScript / Rust) is a small
empirical check on whether the metaprogramming benefit is host-
language-bound or substrate-bound, and lands closer to the
multilingual line than to the model-architecture line.

\paragraph{Class-level and pragmatic code-generation benchmarks.}
ClassEval~\citep{du2024classeval} and
CoderEval~\citep{yu2024codereval} extend execution-based code
evaluation from single-function targets (HumanEval, MBPP, APPS)
to richer real-code structures. We do not use either directly
because both target mainstream languages where the per-language
ecosystem prior is strong; the controlled variable in our setting
is exactly the absence of that prior.

\paragraph{Long-horizon reasoning under static prompting.}
LongCoT~\citep{motwani2026longcot} probes long chain-of-thought
reasoning capability under fixed prompts. Our setting is closer
to long-horizon \emph{tool-use} than long-horizon reasoning per
se: the agent's external state (workspace, generator file, local
test runs) carries most of the long-horizon information, not its
own monologue.

\paragraph{Benchmark validity, robustness, and contamination.}
A line of work warns that high benchmark scores can be brittle:
CheckList~\citep{ribeiro2020checklist} introduces behavioral
testing for NLP models;
Adversarial NLI~\citep{nie2020anli} and
CrossFit~\citep{ye2021crossfit} stress-test against adversarial
or cross-task generalization;
\citet{bowman2021benchmarking} discuss what it would take to fix
NLU benchmarking. On contamination specifically,
\citet{oren2024proving} provide black-box tests for training-set
contamination, \citet{deng2024contamination} investigate
contamination in modern LLM benchmarks, and
\citet{xu2024survey} survey the broader landscape. We do not
claim formal distributional novelty for the four target esolangs;
Appendix~\ref{app:contamination} reports public-code prevalence
and $n$-gram overlap analyses motivated directly by this line.

\paragraph{Mainstream-benchmark sourcing checks.}
For transparency on the mainstream-benchmark rows of
Table~\ref{tab:cliff-spread}, the Vals.ai
SWE-Bench Verified leaderboard~\citep{valsai_swev_2026},
LiveCodeBench v6 leaderboard~\citep{valsai_lcb_2026},
and Terminal-Bench 2.0 leaderboard~\citep{valsai_terminalbench_2026}
were used as third-party verification of vendor-published numbers
where applicable; per-agent attribution is in
Table~\ref{tab:swev-sourcing}.

\paragraph{Cognitive-science framing.}
The view that agents reorganize hard problems by building
external structure, rather than solving them entirely
``in the head,'' has a long history in cognitive science.
\citet{hutchins1995cognition} introduced \emph{distributed
cognition} and \citet{clark1998extended} the \emph{extended mind}
hypothesis. We do not attempt to evaluate these as theories of
LLM cognition. We use the framing only as a label for the
empirical pattern we observe: in our setting, the strongest agents
externalise fragile target-language state into named, reusable
host-language primitives, and the resulting ``scaffolding'' is
itself the locus of capability differences between agents.

\section{Future work}
\label{app:future-work}

Constructing a new programming language designed to be genuinely
out-of-distribution, such as a niche or constructed substrate,
would let us test the gap under fully controlled conditions without
the contamination concerns inherited by public esolangs. Extending
the same methodology to non-code environments where tools and
external state matter (scientific workflows, data analysis, theorem
proving, interactive web tasks) would test whether similar hidden
capability gulfs appear once surface fluency is removed. Benchmark
families that vary surface unfamiliarity, semantic complexity, tool
access, and program length independently would let us attribute the
gap to specific factors rather than treat it as monolithic.
Agent-level diagnostics that measure how local feedback converts
into solves, and metrics for adaptive tool use under unfamiliar
interfaces, would make agentic adaptation measurable in its own
right. A natural follow-up to our reference-library finding is
whether agents can persist accumulated structure as local notes or
textbooks and build on that knowledge across runs, turning a
one-shot library into long-horizon scaffolding. Together these open
a research program on out-of-distribution agentic adaptation,
focused on how agents construct working interfaces to unfamiliar
environments using tools, external structure, and intermediate
representations.

\newpage
\section*{NeurIPS paper checklist}

This checklist reflects the current draft. The anonymous supplementary
archive (harness, interpreters, prompts, all 48 experiment cells, smoke and
rigorous end-to-end tests) is included with this submission and is referred
to throughout this checklist.

\begin{enumerate}
  \item \textbf{Claims.} Do the main claims made in the abstract and
  introduction accurately reflect the paper's contributions and scope?\\
  Answer: Yes.\\
  Justification: The draft states an empirical claim about agent-system
  adaptation under unfamiliar executable interfaces and explicitly avoids a
  formal distributional-novelty claim.

  \item \textbf{Limitations.} Does the paper discuss limitations of the work?\\
  Answer: Yes.\\
  Justification: Section~\ref{sec:limitations} discusses training-data
  uncertainty, the mixture of unfamiliarity and difficulty, cross-protocol
  cells, agent-wrapper effects, and the artificial nature of esolangs.

  \item \textbf{Theory assumptions and proofs.} For each theoretical result,
  does the paper provide assumptions and proofs?\\
  Answer: N/A.\\
  Justification: The paper is empirical and does not claim new theoretical
  results.

  \item \textbf{Experimental result reproducibility.} Does the paper disclose
  information needed to reproduce the main experimental results?\\
  Answer: Yes.\\
  Justification: The methodology and appendix describe the sequential harness,
  problem order, model wrappers, budget regimes, hidden-submission rule, and
  solved-task scoring rule. The accompanying anonymous supplementary archive
  ships the harness, the four esoteric-language interpreters, the per-language
  language-reference prompts (Appendix~\ref{app:system-prompts}), the four
  experiment configurations (main grid, metaprogramming ablation, distillation,
  cross-language transfer) wired as 48 ready-to-run cells, and a smoke test
  plus rigorous end-to-end test that exercise every harness command path
  without requiring any provider API key. The harness emits per-cell JSON
  exports (\texttt{python harness.py export}) that contain the full per-problem
  fetch/run/submit/skip event log used to compute the reported numbers.

  \item \textbf{Open access to data and code.} Does the paper provide open
  access to data and code with reproduction instructions?\\
  Answer: Yes.\\
  Justification: The dataset (EsoLang-Bench) is a previously released
  third-party artifact, publicly hosted at the canonical URL referenced in
  Section~\ref{sec:eval}. The harness, interpreters, prompts, experiment
  scaffolds, and reproducibility scripts are released under the MIT license
  in the anonymous supplementary archive accompanying this submission;
  \texttt{README.md} and \texttt{HOWTO\_RUN.md} document the four reviewer
  paths from a 10-second smoke test (no key) through the rigorous end-to-end
  test (no key) to running an agent against any of the 48 cells (provider key
  required). The archive uses only the Python standard library plus the
  third-party \texttt{shakespearelang} package for the Shakespeare interpreter.

  \item \textbf{Experimental setting/details.} Does the paper specify the
  experimental settings needed to understand the results?\\
  Answer: Yes.\\
  Justification: Section~\ref{sec:eval} of the body and Appendix
  Table~\ref{tab:setup} together specify the primary protocol's
  task substrate, problem order, hidden-test rule, hidden-submission
  cap, local interpreter call regime, per-turn output token budget,
  workspace isolation, sampling settings, aggregation across runs
  (Appendix~\ref{app:robustness}), and uncertainty reporting. Per-condition deviations from the
  primary protocol (no-metaprogramming variant, cross-language
  generator transfer, distillation text-only and library
  conditions, interpreter-budget ablation, token-efficiency
  ablation, controlled-access protocol) are enumerated in
  Appendix~\ref{app:system-prompts} alongside the verbatim
  per-language reference prompts. Per-cell raw counts for every
  reported number are reproducible from the
  \texttt{export.json} files emitted by
  \texttt{python harness.py export} in each of the 48 cells shipped
  in the supplementary archive; the rigorous end-to-end test
  (\texttt{scripts/rigorous\_test.sh}) exercises this export path
  without requiring any provider API key.

  \item \textbf{Experiment statistical significance.} Does the paper report error
  bars or appropriate uncertainty information?\\
  Answer: Yes.\\
  Justification: All four esoteric-language columns in
  Table~\ref{tab:main-results} (Brainfuck, Befunge-98, Whitespace,
  Shakespeare) report cells in percentage-solved form with $\pm$95\%
  binomial Wilson half-widths over 80 problems per language, as
  stated in the table caption. The Reporting paragraph in
  Section~\ref{sec:eval} states that headline cells report the
  Session 1 solved count (with two further independent sessions per
  cell as sanity checks, tabulated in
  Appendix~\ref{app:per-session-results}), and that ablation cells
  are means of two independent sessions. Full per-cell asymmetric
  Wilson $95\%$ confidence intervals are tabulated in
  Appendix~\ref{app:uncertainty}; the session-aggregation protocol is
  in Appendices~\ref{app:reporting} and~\ref{app:robustness}. The same Wilson interval
  treatment carries through to the cross-harness control
  (Appendix~\ref{app:cross-harness}), the metaprogramming ablation
  (Section~\ref{sec:abl-meta}), the distillation cells
  (Section~\ref{sec:distillation}), and the cross-language transfer
  cells reported inline in Section~\ref{sec:abl-meta}; none of these
  report a point estimate without an accompanying interval.

  \item \textbf{Experiments compute resources.} Does the paper provide
  compute-resource information?\\
  Answer: Yes.\\
  Justification: Appendix Table~\ref{tab:setup} specifies the
  per-turn output token budget, the local interpreter call regime,
  the number of hidden submissions per problem, and the sampling
  settings (provider / wrapper defaults). The token-efficiency
  analysis in Section~\ref{sec:resources} reports cumulative API
  output tokens per cell on the easy tier. Local wall-clock and
  operator hardware are not reported because all evaluated agents
  run as managed APIs on vendor infrastructure rather than on local
  accelerators, so compute is fully described by the per-cell API
  output-token budget plus the interpreter-call regime above.

  \item \textbf{Code of ethics.} Does the research conform to the NeurIPS Code of
  Ethics?\\
  Answer: Yes.\\
  Justification: The work evaluates existing coding agents on programming tasks
  and does not involve human-subject experiments, private personal data, or model
  training.

  \item \textbf{Broader impacts.} Does the paper discuss positive and negative
  societal impacts?\\
  Answer: Yes.\\
  Justification: The evaluation reported here is intended to improve
  the visibility of capability differences among coding agents in
  long-tail settings relevant to real deployments. Positive impact:
  clearer evaluations in low-ecosystem programming environments help
  practitioners pick and budget agents for internal DSL work, legacy
  integration, and closed-source platform development, where current
  leaderboards are poorly predictive. Negative impact: the same
  evaluations can be misused as ranking claims that extend beyond the
  measured regime; we mitigate this by reporting the primary protocol
  and the controlled-access protocol separately and flagging
  incomplete cells explicitly. No new model, weights, or dataset
  capable of offensive use is released in this work, and all
  evaluated systems are already publicly available.

  \item \textbf{Safeguards.} Does the paper describe safeguards for responsible
  release of high-risk assets?\\
  Answer: N/A.\\
  Justification: The paper evaluates existing models and benchmark harnesses
  rather than releasing a model, exploit, or high-risk dataset.

  \item \textbf{Licenses for existing assets.} Are existing assets credited and
  licenses respected?\\
  Answer: Yes.\\
  Justification: The EsoLang-Bench
  dataset~\citep{esolangbench2026} is a previously released
  third-party artifact, cited at its canonical public URL with the
  dataset paper referenced; we consume it without modification under
  its public release terms. Mainstream-benchmark scores re-used from
  public vendor reports (SWE-Bench Verified, Terminal-Bench 2.0,
  LiveCodeBench v6) are sourced per agent in
  Appendix Table~\ref{tab:swev-sourcing}, with each row attributing
  the cited Anthropic system card / OpenAI system card / Moonshot
  release / Vals.ai third-party
  leaderboard~\citep{anthropic_models_2026,openai_gpt54_2026,openai_gpt54mini_2026,moonshot_kimi_k25_2026,valsai_swev_2026,valsai_lcb_2026,valsai_terminalbench_2026}
  as the source. This submission uses the official NeurIPS 2026
  style file without modification. Code shipped in the supplementary
  archive (harness, four esoteric-language interpreters, prompts,
  experiment scaffolds, reproducibility scripts, distillation
  reference library) is released under the MIT license; the
  Shakespeare interpreter wraps the third-party
  \texttt{shakespearelang} package, used unmodified under its own
  license.

  \item \textbf{New assets.} Are new assets introduced in the paper documented?\\
  Answer: Yes.\\
  Justification: The new assets introduced by this paper are
  (i) the evaluation harness wrapping EsoLang-Bench (released in the
  supplementary archive under \texttt{benchmark\_harness/} with
  \texttt{README.md} and \texttt{HOWTO\_RUN.md} as entry points);
  (ii) the four per-language language-reference prompts
  (\texttt{prompts/<lang>/CLAUDE.md} and \texttt{AGENTS.md}, identical
  content); (iii) the four experiment configurations as 48
  ready-to-run cells under \texttt{experiments/}; and (iv) the
  reference library scaffolds for the distillation condition under
  \texttt{experiments/03\_distillation/reference\_lib/}. All are
  documented in their respective \texttt{README.md} files. Trace
  excerpts reproduced in Appendix~\ref{app:trace-examples} are
  selected by the rule stated in Appendix~\ref{app:trace-examples}
  (single recorded session, four pre-specified phenomena).

  \item \textbf{Crowdsourcing and human subjects.} Does the paper include details
  for crowdsourcing or human-subject work?\\
  Answer: N/A.\\
  Justification: The work does not involve crowdsourcing or human-subject
  experiments.

  \item \textbf{IRB approvals.} Does the paper describe IRB approvals or
  equivalent review for human-subject work?\\
  Answer: N/A.\\
  Justification: The work does not involve human-subject experiments.

  \item \textbf{Declaration of LLM usage.} Does the paper describe LLM usage when
  it is part of the core method?\\
  Answer: Yes.\\
  Justification: The evaluated systems are LLM-based coding agents; the
  methodology section describes model snapshots, agent wrappers, and harness
  interaction.
\end{enumerate}

\end{document}